\documentclass{article}

\usepackage{PRIMEarxiv}

\usepackage[utf8]{inputenc} 
\usepackage[T1]{fontenc}    
\usepackage{hyperref}       
\usepackage{url}            
\usepackage{booktabs}       
\usepackage{amsfonts}       
\usepackage{nicefrac}       
\usepackage{microtype}      
\usepackage{lipsum}
\usepackage{fancyhdr}       
\usepackage{graphicx}       
\graphicspath{{media/}}     

\title{Controlled physics-informed data generation for deep learning-based remaining useful life prediction under unseen operation conditions
\thanks{\textit{Citation}: 
J. Xiong, O. Fink, J. Zhou, and Y. Ma, Controlled physics-informed data generation for deep learning-based remaining useful life prediction under unseen operation conditions. Mech. Syst. Sig. Process., 2023. 197. DOI: 10.1016/j.ymssp.2023.110359.} 
}

\author{
  Jiawei Xiong \\
  School of Economics and Management \\
  Nanjing University of Science and Technology \\
  Nanjing, 210094 \\
  China \\
  \texttt{hsiungchiawei@njust.edu.cn} \\
  \And
  Olga Fink \\
  Laboratory of Intelligent Maintenance and\\
  Operation Systems, EPFL \\
  1015 Lausanne\\
  Switzerland \\
  \texttt{olga.fink@epfl.ch} \\
  \And
  Jian Zhou \\
  School of Economics and Management \\
  Nanjing University of Science and Technology \\
  Nanjing, 210094 \\
  China \\
  \texttt{zhoujian977913@njust.edu.cn} \\
  \And
  Yizhong Ma\thanks{Corresponding author.} \\
  School of Economics and Management \\
  Nanjing University of Science and Technology \\
  Nanjing, 210094 \\
  China \\
  \texttt{yzma@njust.edu.cn} \\
}

\begin{document}
\maketitle

\begin{abstract}
Limited availability of representative time-to-failure (TTF) trajectories either limits the performance of deep learning (DL)-based approaches on remaining useful life (RUL) prediction in practice or even precludes their application. Generating synthetic data that is physically plausible is a promising way to tackle this challenge. In this study, a novel hybrid framework combining the controlled physics-informed data generation approach with a deep learning-based prediction model for prognostics is proposed. In the proposed framework, a new controlled physics-informed generative adversarial network (CPI-GAN) is developed to generate synthetic degradation trajectories that are physically interpretable and diverse. Five basic physics constraints are proposed as the controllable settings in the generator. A physics-informed loss function with penalty is designed as the regularization term, which ensures that the changing trend of system health state recorded in the synthetic data is consistent with the underlying physical laws. Then, the generated synthetic data is used as input of the DL-based prediction model to obtain the RUL estimations. The proposed framework is evaluated based on new Commercial Modular Aero-Propulsion System Simulation (N-CMAPSS), a turbofan engine prognostics dataset where a limited avail-ability of TTF trajectories is assumed. The experimental results demonstrate that the proposed framework is able to generate synthetic TTF trajectories that are consistent with underlying degradation trends. The generated trajectories enable to significantly improve the accuracy of RUL predictions.
\end{abstract}


\section{Introduction}
In the field of Prognostics \& Health Management (PHM), prognostics refers to the prediction of the remaining useful life (RUL) of an industrial asset \cite{ellefsen2019remaining}. Accurate RUL predictions with sufficiently long foresight periods enable decision-makers to plan the maintenance actions in advance and forestall failures. It not only helps to reduce system downtime, but also improves the cost efficiency \cite{zio2022prognostics}. With the increasing availability of condition monitoring (CM) data for industrial assets, data-driven approaches for predicting RUL have recently attracted the attention of both practitioners and researchers.

The evolution of the failure conditions of an asset is usually affected by different factors, including the environmental and operating conditions. Several approaches have been proposed for predicting the RUL of complex assets \cite{yang2019remaining, da2020remaining}. The three main types of approaches for predicting RUL are model-based approaches \cite{huang2017remaining, qiu2002damage}, data-driven approaches \cite{zhu2018estimation, song2019generic}, and hybrid approaches that combine both model-based and data-driven approaches \cite{chao2022fusing, russell2022physics}. Model-based approaches usually create a system model according to the in-depth understanding of the underlying physics-of-failure processes \cite{nguyen2023physics}. While these approaches are usually effective, they may be computationally intensive to be applied in real time. Moreover, they are often only developed for rather simple systems with only few well-understood failure modes. Even in such cases, the developed models may not be able to fully represent all the underlying failure evolution processes. In addition, since their development is time-consuming and expensive, detailed physics-of-failure models are only available for expensive and safety-critical models. All these limitations hinder the widespread applications of physics-based prognostics approaches.

Data-driven approaches, including statistical and machine learning (ML) methods, do not require a detailed modelling of the underlying physics-of-failure mechanisms. However, they rely on the availability of large representative datasets with time-to-failure (TTF) trajectories that capture the different influencing factors of fault evolution, such as operating and environmental conditions. Different types of neural networks have been pro-posed for prognostics, including convolutional neural networks (CNNs) \cite{zhu2018estimation, zhou2021autonomous}, Generative adversarial networks (GANs) \cite{liu2019simultaneous}, deep belief networks (DBNs) \cite{kim2020bayesian} and recurrent neural networks (RNNs) \cite{shi2021dual, yu2020improved}. Since failures in safety-critical systems are rare, the applicability of data-driven approaches in real applications is limited. While several research studies have proposed different data-driven prognostics approaches. Their performance has typically been either evaluated on non-realistic simulated data (e.g., the old CMAPSS dataset \cite{saxena2008damage}) or on not openly available datasets which precludes the comparison of any newly developed approaches. These ap-proaches may not perform similarly well when applied to other case studies under more realistic conditions.

Data-driven prognostics approaches are typically defined as a regression problem between sensor measurements and real RUL values or the corresponding health indicator. In other words, a regression relationship needs to be developed while taking the influencing factors into consideration. Since the operating conditions (OCs) may be very diverse, a sufficiently large representative training dataset is required to learn the corresponding relationships. In some cases, dedicated regression models may need to be developed for different failure modes (FMs). Since the datasets need to cover all the relevant operating and environmental conditions and since faults are rare and the data collection may not cover the entire trajectories, obtaining enough representative TTF trajectories is still difficult in practice. One of the additional challenges preventing the collection of full TTF trajectories is preventive maintenance which has been widely used for safety-critical systems and prevents the collection of data in the last stage of the fault evolution. If the available training dataset is not fully representative, one possible solution is using transfer learning (TL) \cite{da2020remaining, yang2022multi}. It can improve the prediction performance by learning the information from other similar datasets and then adapt the models based on a limited number of TTF trajectories to the specific conditions of the considered case study. However, these methods can only be applied if labeled datasets collected from systems that are sufficiently similar to the ones considered are available. This requirement may be difficult to meet for safety-critical systems. Another promising solution is generating synthetic TTF. This type of methods is usually effective in scenarios where real representative data is not limited, and there is a pressing need to obtain the data belonging to the unseen categories. Their goal is to generate synthetic data through learning the distribution of the real CM data and then sampling from the learned distribution. This type of approaches usually can be divided into two categories, 1) generators that learn the distributions from real data, and 2) generators based on user-defined distributions combined with Monte Carlo sampling \cite{desai2021timevae}. For the first type, the most frequently applied methods include generative adversarial networks (GANs) \cite{yoon2019time, zhao2022new, li2022tts} and variational autoencoders (VAEs) \cite{desai2021timevae, chen2021trajvae} The main advantage of this type of the generators is that they can automatically learn the distributions of real data and generate synthetic data that follows these distributions. However, they require a large amount of real data, and their training process may be time-consuming. Moreover, for engineered systems, it is often difficult to verify the consistency of the generated samples with the underlying physical laws and the generated data may, therefore, not be interpretable. The second type of methods is typically interpretable, they allow the user to control data generation and integrate domain knowledge \cite{beiden2003general}. Their main limitation, however, is that the synthetic data may differ significantly from the real data since the distributions are assumed and not learned from data.

While they originated in computer vision applications, GANs and VAEs have recently also been increasingly applied in generating time series \cite{li2017adversarial, ramponi2018t}. Several approaches have been proposed for multi-variate time series generation. Esteban \textit{et al.} \cite{esteban2017real} attempted to generate medical time series by replacing CNN with long short-term memory (LSTM) in the original GAN framework to capture long-term temporal correlation. Yoon \textit{et al.} \cite{yoon2019time} proposed a novel framework called Time series Generative Adversarial Networks (TimeGAN) for generating realistic time series data. It combines the flexibility of the unsupervised paradigm with the control provided by supervised training, which demonstrated good generation ability. However, the abovementioned methods are not designed for time series generation for complex engineered systems that does not only generate realistic trajectories but also enables to control the time series generation based on the underlying physical processes, such as degradation levels, fault severities or specific operating conditions. These requirements cannot be satisfied if standard GAN or VAE architectures are applied in PHM directly.

Because of the excellent performance of GANs, recently, they have been applied in several PHM applications. In fault diagnosis, to solve the problem of unbalance data in DL-based diagnosis, Zhou \textit{et al.} \cite{zhou2020deep} proposed a GAN-based approach with new generator and discriminator to generate more synthetic fault samples to improve classification accuracy. Liu \textit{et al.} \cite{liu2022imbalanced} proposed an improved multi-scale residual GAN (MsR-GAN) with feature enhancement-driven capsule network (CapsNet) to generate high-quality fake time-frequency features to balance fault data distribution. To obtain a better transferability between two different domains where only the healthy data class is shared, Rombach \textit{et al.} \cite{rombach2023controlled} proposed a controlled data generation framework based on Wasserstein GAN to generate previously unobserved target faults with different severity levels. In prognostics, considering limited failure data availability, He \textit{et al.} \cite{he2022semi} developed a semi-supervised GAN regression model for RUL predictions. Instead of simply treating suspension histories as unlabeled data, GAN used conditional multi-task objective functions to capture useful information from suspension histories to improve prediction accuracy. Most of these GAN-based methods are used to generate new data by learning to capture the joint probability to improve the accuracy of the model, while they do not consider the physical plausibility of the generation of TTF trajectories under unseen operation conditions, which are important for real-world PHM applications.

Hybrid approaches combine the benefits of both, model-based and data-driven approaches. This type of approaches has the potential to obtain performance improvement by leveraging the advantages of each approach \cite{chao2019hybrid}. Raissi \textit{et al.} \cite{raissi2019physics} proposed physics-informed neural networks (PINNs), a type of neural networks that are trained from additional information obtained by the physical laws for solving supervised tasks. Moreover, specialized network architectures may be designed to satisfy some of the physical invariants automatically for better accuracy, faster training and improved generalization \cite{karniadakis2021physics}. In order to make the DL-based model physically meaningful, Shen \textit{et al.} \cite{shen2021physics} proposed a hybrid approach for bearing fault detection. This model combines a threshold model that obtained from physical knowledge of bearing faults and a deep CNN model. Yucesan \textit{et al.} \cite{yucesan2022hybrid} proposed a new hybrid model to quantify the grease damage of bearing fatigue, where a reduced-order physical model that is designed to accumulate of bearing fatigue damage is embedded in a RNN cell. Yan \textit{et al.} \cite{yan2022integration} proposed an approach that effectively integrate physics-based signal processing technologies and an interpretable artificial network structure for machine degradation modeling. The superiority of this approach is full transparency and interpretability. Arias Chao \textit{et al.} \cite{chao2022fusing} proposed a hybrid architecture that can infer the unobserved process variables for a physics-based model to enhance the input space of a data-driven diagnostics model. However, few existing hybrid approaches have considered the introduction of physics-consistent regularizer in the time-series generation process in industrial prognostics. Thus, a potential direction to ensure that the generated trajectories are following the underlying physical laws is to impose physics-based regularization and specific constraints in the generation process.

In this work, we focus on the challenging problem of RUL prediction when the available TTF trajectories are not sufficiently representative, more specifically when the operating conditions observed in the testing dataset are not represented in the training dataset. The proposed physics-informed generative framework for prognostics enables on the one hand to control the generation of the TTF trajectories and ensures on the other hand that the generated trajectories are realistic, are fully interpretable and are consistent with the underlying degradation processes. We refer to the proposed framework as controlled physics-informed GAN (CPI-GAN). The proposed framework has two main contributions: First, a controlled generator is developed to combine GAN and RNN to ensure that the generator can capture long-term dependencies that are of particular relevance for prognostics applications. Moreover, domain knowledge and physical laws are used as the basic physics constraints during the generation of synthetic TTF trajectories. Second, a physics-informed loss function based on system health indicators inferred by a surrogate model is proposed to ensure that the synthetic data is consistent with the true fault evolution trajectory. The interactions between the generator, the constraints, and the physics-informed loss function help to ensure that the synthetic data are consistent with the real underlying physical laws.

The remainder of the paper is organized as follows. In Section \ref{sec:Background}, the background of the methodology and the general definition of prognostics problem are introduced. Section \ref{sec:Methodology} presents the details of the proposed CPI-GAN framework. In Section \ref{sec:Case study} the case study is introduced. In Section \ref{sec:Results and discussion}, the obtained results are presented and discussed. Finally, in Section \ref{sec:Conclusions}, the conclusion of this study is presented.

\section{Background}
\label{sec:Background}
This section introduces the integral parts of the proposed framework: the basic concepts, the notations of the system performance model, the system degradation model, and the general formulation of the prognostics problem.

\subsection{System performance model}
For a complex engineered system, system performance is usually modeled by high-fidelity physical models. These system performance models can output measurable process variables and system performance that cannot be directly measured. Because complex systems are often coupled by multiple components, system performance models cannot be described using explicit equations, so they are often represented mathematically as nonlinear equations. In this research, we consider the turbofan engine system simulated by the new Commercial Modular Aero-Propulsion System Simulation (N-CMAPSS) \cite{arias2021aircraft} as an example. The inputs of the performance model include scenario-descriptor operating conditions ($w$) and the unobservable model health parameters ($\theta$).

Where health parameters ($\theta$) are model tuners that are used to compensate the deteriorated behavior of the system (i.e., component efficiencies, flow). In N-CMAPSS, five rotating sub-components can be affected by the degradation in flow and efficiency. Concretely, these systems are fan, low-pressure compressor (LPC), high-pressure compressor (HPC), low-pressure turbine (LPT) and high-pressure turbine (HPT). Depending on the fault type, different health parameters will be affected. The outputs of the system performance model are estimates of the measured physical properties   and unobserved properties   (i.e., virtual sensors). Typically, the nonlinear performance model can be expressed as:
\begin{equation}
[{x_s}^{(t)},{x_v}^{(t)}] = {\cal F}({w^{(t)}},{\theta ^{(t)}})
\end{equation}

\subsection{System degradation model}
In N-CMAPSS, the degradation process of each engine is divided into three steps: an initial degradation, a normal degradation and abnormal degradation. The failure modes are mainly manifested by a continuous degradation of the engine rotating sub-components.

\paragraph{Initial degradation.}
The first step of the degradation is caused by manufacturing and assembly tolerances. The initial degradation of each engine’s sub-component differs typically only slightly. 

\paragraph{Normal degradation.}
The second step of the degradation of each engine’s sub-component is caused by wear and tear resulting from long-term usage. 

\paragraph{Transition from normal to abnormal degradation.}
At a point in an engine’s life cycle, ${t_s}$, a particular failure mode might cause engine’s health state transition from normal degradation to an abnormal state, and lead to an eventual failure at ${t_{{\rm{EOL}}}}$ (i.e., end-of-life).

\paragraph{Abnormal degradation.}
The abnormal degradation starts with the fault initiation (e.g., ${t_s}$) and ends when the acceptable level of per-formance cannot be reached. The detailed model of the evolution of the abnormal system degradation with time can be found in \cite{saxena2008damage}.

\subsection{General formulation of a prognostics problem}
Given a set of multivariate time-series of CM sensors readings ${X_{{s_i}}} = {[x_{{s_i}}^{(1)},...,{\rm{ }}x_{{s_i}}^{({m_i})}]^T}$ and their corresponding RUL true labels, i.e., ${Y_i} = {[y_i^{(1)},...,{\rm{ }}y_i^{({m_i})}]^T}$ from a fleet of $U$ units, $i \in \left( {1,{\rm{ }}...{\rm{ , }}U} \right)$. Observation at each time step $x_{{s_i}}^{(t)} \in {R^p}$ is a vector of $p$ real measurements taken at operating conditions $w_i^{(t)} \in {R^s}$. Where ${m_i}$ is the length of the sensory signal for the $i{\rm{ - th}}$ unit. The length can be different from unit to unit in general. The total combined length of the dataset is $m = \sum\nolimits_{i = 1}^U {{m_i}} $. Thus, the training dataset can be expressed as  ${{\cal D}_{{\rm{real}}}} = \{ {W_i},{X_{{s_i}}},{Y_i}\} _{i = 1}^U$. The systems usually start from an unknown initial health condition, the degradation process of the system’s com-ponents is recorded in the CM data of each unit. The degradation process of system’s components experiences two stages in general. In the first stage, components experience normal degradation until point in time ${t_{{s_i}}}$, when a fault is induced and an abnormal condition starts, leading to systems cannot be able to fulfil the defined requirements at ${t_{{\rm{EO}}{{\rm{L}}_i}}}$.

Given this set-up, in the first step, we develop a controlled generation model based on the real data ${{\cal D}_{{\rm{real}}}}$, ${\theta _{{\rm{real}}}}$ and domain knowledge to generate a set of synthetic data ${{\cal D}_{{\rm{syn}}}} = \{ {W_g},{X_{{s_g}}},{Y_g}\} _{g = 1}^G$ of $G$ units, $g \in \left( {1,{\rm{ }}...{\rm{ , }}G} \right)$, where ${X_{{s_g}}} = {[x_{{s_g}}^{(1)},...,{\rm{ }}x_{{s_g}}^{({h_g})}]^T}$ are synthetic multivariate time-series of CM sensors readings generated at operating conditions $w_g^{(t)} \in {R^s}$ that are different from $w_i^{(t)}$. The total combined length of the synthetic dataset is ${m_{{\rm{syn}}}} = \sum\nolimits_{g = 1}^G {{h_g}} $. Then, the synthetic data ${{\cal D}_{{\rm{syn}}}}$ is used to complement the available data space of the real training data ${{\cal D}_{{\rm{real}}}}$. More compactly, we denote the enhanced dataset as ${{\cal D}_{{\rm{real + syn}}}} = \{ {W_i},{X_{{s_i}}},{Y_i}\} _{i = 1}^{U + G}$. In the second step, we develop a prognostic model ${\cal G}{_H}$ from ${{\cal D}_{{\rm{real + syn}}}}$ that can provide reliable RUL estimations ${\hat Y_j} = [{\hat y_j}^1,...,{\rm{ }}{\hat y_j}^{{k_j}}]$ on a testing dataset of $V$ units ${{\cal D}_{{\rm{test}}}} = \{ {W_j},{X_{{s_j}}}\} _{j = 1}^V$, $j \in \left( {1,{\rm{ }}...{\rm{ , }}V} \right)$, where ${X_{{s_j}}} = {[x_{{s_j}}^{(1)},...,{\rm{ }}x_{{s_j}}^{({k_j})}]^T}$ are multivariate time-series of CM sensors readings taken at operating conditions $w_j^{(t)} \in {R^s}$. The total combined length of the testing dataset is ${m_*} = \sum\nolimits_{j = 1}^V {{k_j}} $.

The performance of the developed model is evaluated using the root-mean-square error (RMSE), NASA’s scoring function $score$ \cite{saxena2008damage}, and the prediction horizon \cite{chao2022fusing} with a prediction error below 5 cycles (${H_{\left| {{\Delta _c}} \right| \le 5}}$), calculated from the true RUL values ${y_j}$ and the RUL estimations ${\hat y_j}$. Three performance metrics are shown as follows:

\begin{equation}
RMSE{\rm{  =  }}\sqrt {\frac{1}{{{m_*}}}\sum\limits_{j = 1}^{{m_*}} {{{({\Delta _j})}^2}} }
\end{equation}

\begin{equation}
score = \sum\limits_{j = 1}^{{m_*}} {\exp \left( {\alpha \left| {{\Delta _j}} \right|} \right) - 1}
\end{equation}

\begin{equation}
{H_{\left| {{\Delta _c}} \right| \le 5}} = {t_{EOL}} - {t_{\left| {{\Delta _c}} \right| \le 5}}
\end{equation}

where ${m_*}$ denotes the total number of testing samples, ${\Delta _j}$ is the difference between the true and the predicted RUL for the $j{\rm{ - th}}$ sample (${\Delta _j} = {y_j} - {\hat y_j}$), and $\alpha $ is ${1 \mathord{\left/
 {\vphantom {1 {13}}} \right.
 \kern-\nulldelimiterspace} {13}}$ if RUL is under-estimated and ${1 \mathord{\left/
 {\vphantom {1 {10}}} \right.
 \kern-\nulldelimiterspace} {10}}$ if it is overesti-mated. Hence, the scoring function is asymmetric and penalizes over-estimations of the RUL. ${t_{\left| {{\Delta _c}} \right| \le 5}}$ is the cycle time in which the prediction error is below 5 cycles for any future time.

\section{Methodology}
\label{sec:Methodology}
\begin{figure}
  \centering
  \includegraphics[scale=0.52]{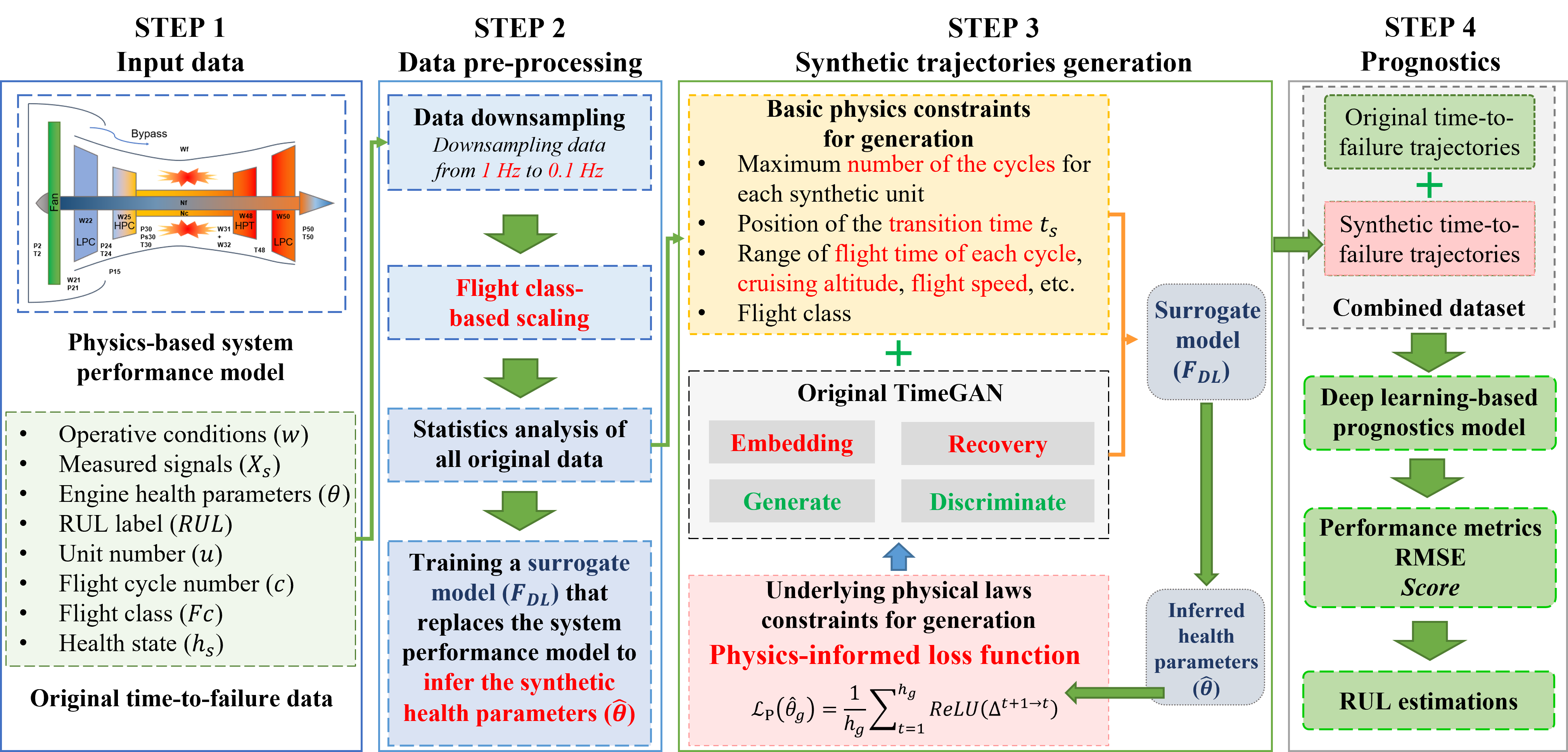}
  \caption{The framework of the proposed controlled physics-informed data generation for RUL prediction.}
  \label{fig:fig1}
\end{figure}

In this section, the details of the proposed synthetic data generation approach and the DL-based prognostics model are introduced. An overview of the proposed hybrid prediction framework is shown in Figure \ref{fig:fig1}.

The proposed framework includes four steps: input data, data pre-processing, synthetic trajectories generation and prognostics. The first step is to select the variables needed to construct the hybrid model. The second step is to pre-process the data, including downsampling (reducing the magnitude of the data), normalization by flight class (eliminating the effect of different flight classes on the scale of the data), statistical analysis of all available data (obtaining the physical characteristics of the degradation trajectories under various flight classes as basic physics constraints for generation). We assume that during training, we have access to real $\theta $, but not during testing. Therefore, a surrogate model is developed to replace the traditional physics-based system performance model (inferring of the system health state parameters that are not directly observable). In the third step, the obtained basic physics constraints are used as the initial settings of the generator, and after each generation of synthetic sequence, this synthetic sequence is inputted to the surrogate model to infer unobservable health parameters, after which the inferred health parameters are used to impose in the generation as penalty targets of the physics-informed loss function, thus, ensuring that the synthetic sequence is conform with the underlying degradation characteristics of the engine. In the last step, the synthetic degradation trajectories and the original degradation trajectories are combined to a training dataset and used as input to the prognostics model to output the estimations of RUL. The details of the proposed framework are described in detail in the later section.

\subsection{Controlled physics-informed time-series generative adversarial networks}
For complex engineered systems, the two main requirements for synthetic data generation are: 1) synthetic data needs to maintain the consistency of physics with the target data and is consistent with local changes over time (i.e., basic physics constraints); 2) the relationships between synthetic data’s variables are plausible for these operating regimes and the underlying physical laws (i.e., underlying physical laws constraints). The original TimeGAN \cite{yoon2019time} can satisfy the first requirement. However, this approach does not consider the objective of physical constraints when generating condition monitoring data of industrial assets. Thus, in this research, we propose to extend the original TimeGAN by imposing constraints and physics-informed loss functions to ensure that the extended generator can satisfy both of the above requirements. We refer to the new architecture as CPI-GAN.

\begin{figure}
  \centering
  \includegraphics[scale=0.53]{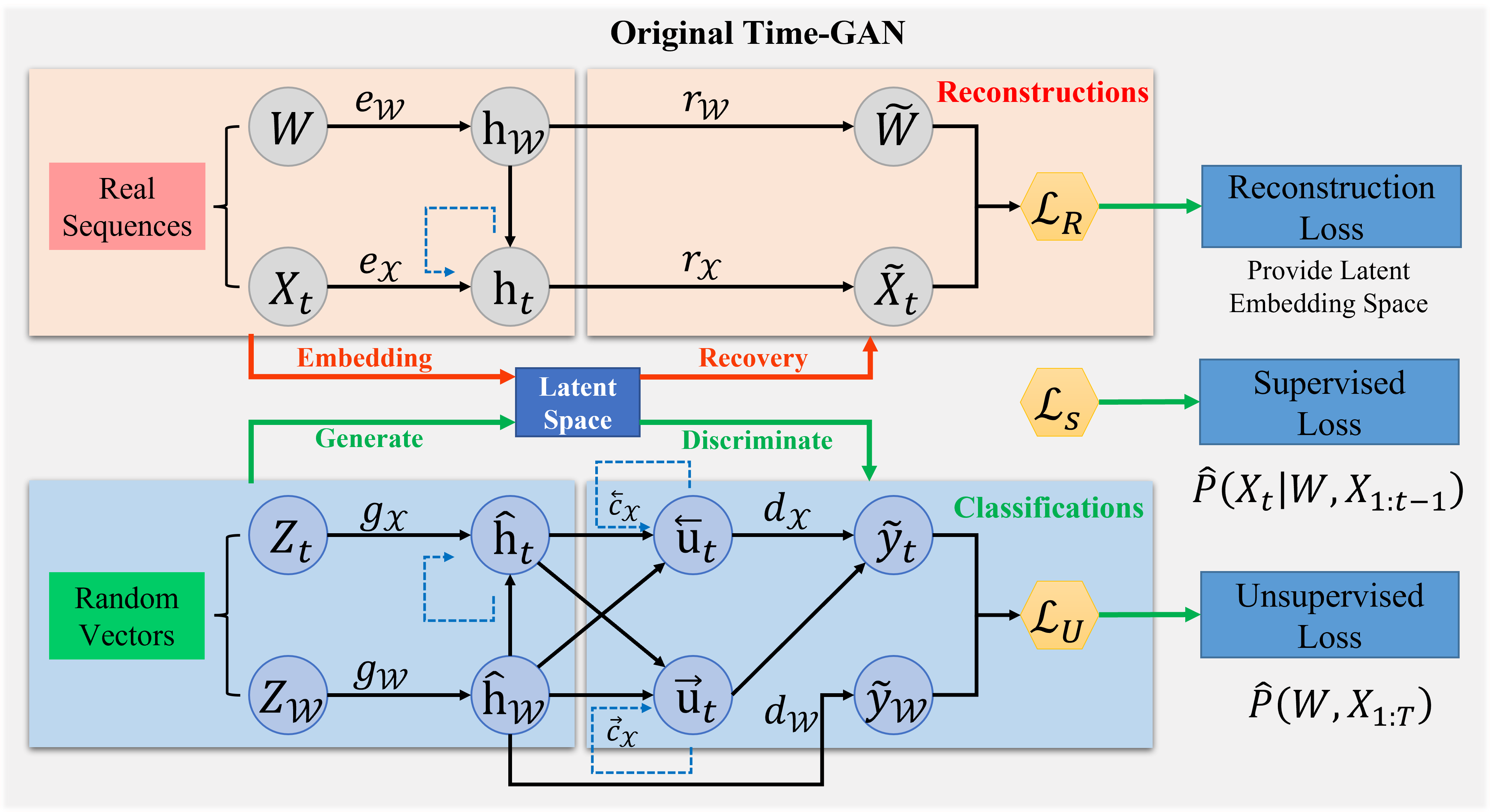}
  \caption{The structure of original TimeGAN (modified from \cite{yoon2019time}).}
  \label{fig:fig2}
\end{figure}

TimeGAN is a universal generation framework for generating realistic time series data in various domains \cite{yoon2019time}. It can generate time-series of any length according to the user's requirements. The original TimeGAN consists of four components: embedding module, recovery module, sequence generator module, and sequence discriminator module. The structure of the original TimeGAN is shown in Figure \ref{fig:fig2} The embedding and recovery modules can represent the time-series as lower dimensional features which reduce the high-dimensionality of the GAN’s learning space. Autoencoding components are trained jointly with the adversarial components, such that TimeGAN can learn to encode features and generate new samples at the same time. The adversarial network operates within the latent space provided by the embedding network. The latent dynamics of both real and synthetic samples are synchronized through a supervised loss.

As shown in Figure \ref{fig:fig2}, the original TimeGAN consists of three loss functions: reconstruction loss, unsupervised loss, and supervised loss. These three loss functions ensure that the synthetic samples have an approximate distribution to the real data both locally and globally. The original TimeGAN performs well on time series data that does not need to comply with the underlying physical processes. However, in complex engineered systems with multiple coupled components, it is not sufficient for synthetically generated samples to comply with the approximated distribution of features. The generated samples must comply with the underlying physical laws. More technical details about TimeGAN can be found in \cite{yoon2019time}. 

\begin{figure}
  \centering
  \includegraphics[scale=0.53]{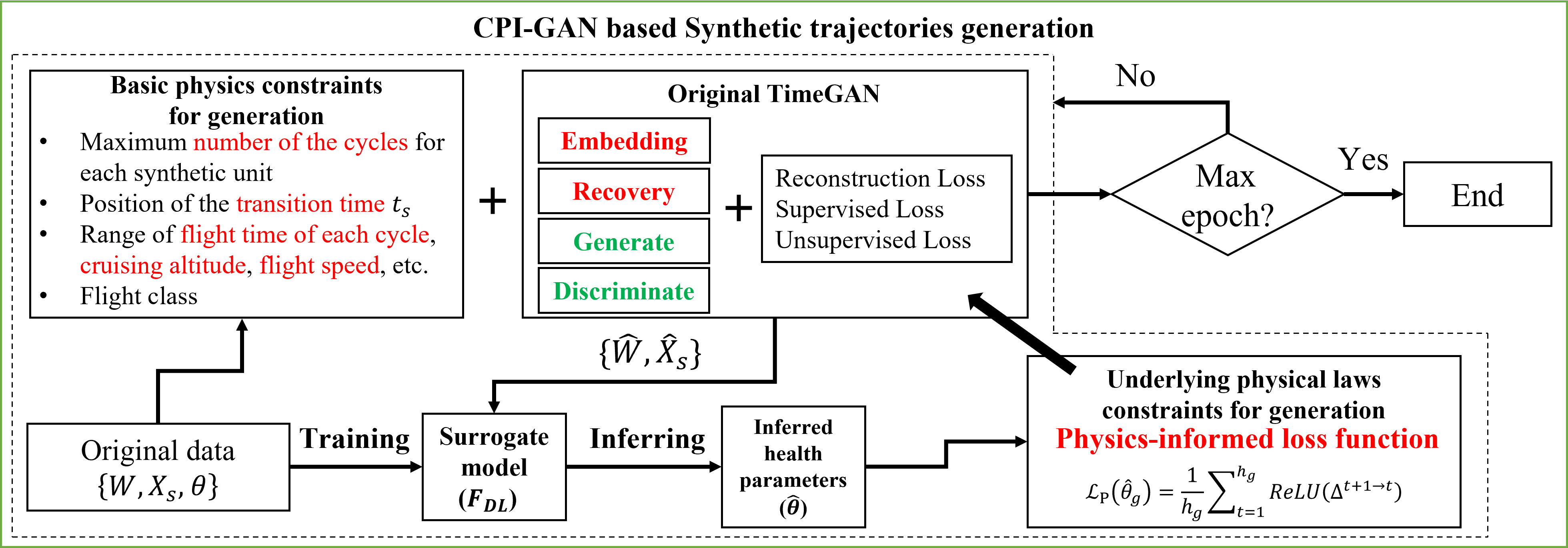}
  \caption{The structure of the CPI-GAN.}
  \label{fig:fig3}
\end{figure}

Industrial assets often operate in complex and changing environments and often operated in discrete cycles, whereby the length of the cycle can vary depending on the mission and the associated operating conditions. Taking turbofan engines as an example, the flight time, altitude, and cruise speed of flights vary significantly depending on the type of flight (e.g., long, or short flights) and the environmental conditions. To ensure that the synthetic data can provide new information to the predictive model, the generated data needs to represent operational conditions and flight class that do not collected in the training set but exist in reality. These generated TTF degradation trajectories that are not covered by the training set also need to obey the simple constraints corresponding to the flight type they belong to (e.g., different flight types correspond to different flight time, altitude, cruise speed, max RUL and the position of  ), and underlying physical laws constraints (e.g., the degradation in flow and efficiency follow the realistic degradation patterns). The structure of the CPI-GAN is shown in Figure \ref{fig:fig3}.

For the basic physics constraints, it is only necessary to determine the range of values for each constraint in the initial stage of CPI-GAN based on the a priori domain knowledge of the target scenario from statistical data analysis. Since CPI-GAN can generate synthetic data of arbitrary length, imposing these constraints based on physical information does not require any changes to the structure from the original generator, only the desired range of values needs to be set in the initial settings of the generator. This allows for controlled synthetic data to be obtained. The procedure of deriving the range of values for these constraints is described in detail in section \ref{sec:Domain knowledge-based basic physics constraints}. For the underlying physical laws constraints, a physics-informed loss function is proposed in CPI-GAN to impose a penalty on the unrealistic part of generation process. The physics-informed loss function is described in detail in the next section.

\subsection{Physics-informed loss function}
In order to ensure that the TTF trajectories are following realistic degradation processes. As mentioned above, the degradation effects are modelled by adjustments of flow capacity and efficiency of these engine sub-components (e.g., the health parameters corresponding to the synthetic trajectories). Therefore, we can en-sure that the degradation process of the synthetic trajectories satisfies the underlying physical laws constraints by penalizing $\theta $ that is not conform with the physical laws in generation process. According to the dataset description document \cite{arias2021aircraft}, the loss in efficiency or flow (e.g., $\theta $) should be globally monotonic throughout the period of degradation. Imposing this as a penalty in the learning process ensures that the values of $\theta $ at the later time step are smaller than the values at the previous time step. 

Considering a set of synthetic TTF trajectories, ${{\cal D}_{{\rm{syn}}}} = \{ {W_g},{X_{{s_g}}}\} _{g = 1}^G$, for any pair of consecutive time steps, $t$ and $t + 1$, the health indicators of $g{\rm{ - th}}$ unit, are related to each other in the following manner:

\begin{equation}
{\Delta ^{t + 1 \to t}} = \theta _g^{t + 1} - \theta _g^t \le 0{\rm{, \quad for  }} \quad t < t + 1, \enspace {\rm{ }}t \in [1,...,{h_g}]
\end{equation}

where ${\Delta ^{t + 1 \to t}}$ denotes the difference between two adjacent time points, ${h_g}$ is the length of the TTF trajectory of unit $g$. At time $t + 1$, if the changing trend of the health parameter, $\theta $, is not following a realistic degradation trend, a penalty will be imposed. In case the generated degradation trend is consistent with the real degradation trend, the loss will be zero. To ensure this property, we apply the rectifier activation function, also named the rectified linear unit (ReLU). The ReLU function clips each value at a threshold ${\lambda _{{\rm{ReLU}}}}$ is expressed as

\begin{equation}
{\rm{ReLU }}\left( {{\Delta ^{t + 1 \to t}}} \right) = \max \left( {0,{\rm{ }}{\Delta ^{t + 1 \to t}}} \right) = \left\{ \begin{array}{l}
{\Delta ^{t + 1 \to t}} \quad {\rm{    }}{\Delta ^{t + 1 \to t}} > {\rm{ }}{\lambda _{{\rm{ReLU}}}}\\
0 \quad \quad \quad \quad {\rm{            }}{\Delta ^{t + 1 \to t}} \le {\rm{ }}{\lambda _{{\rm{ReLU}}}}
\end{array} \right.
\end{equation}

where ReLU ( ) denotes the rectified linear unit and ${\lambda _{{\rm{ReLU}}}}$ denotes the penalty threshold. The threshold can be changed to control the scale of the penalty to be applied. For example, if it is necessary to make the inferred health parameter values maintain an absolute monotonic downward trend, the penalty threshold ${\lambda _{{\rm{ReLU}}}}$ is set to zero. Different threshold values other than zero can be used to impose a different level of penalty based on the practical generation requirements. The selection of the penalty threshold is discussed in detail in section \ref{sec:Sensitivity analysis of penalty threshold}.

If the health parameter variation of $g{\rm{ - th}}$ unit between two subsequent time steps $t$ and $t + 1$, ${\Delta ^{t + 1 \to t}}$, is greater than the penalty threshold ${\lambda _{{\rm{ReLU}}}}$ ( which can be viewed as a violation of Eq. (6)), the penalty is applied. When $t \in [1,...,{h_g}]$, the mean of all penalties across every consecutive time-step is used as the final physics-informed loss function, the final physics-informed loss function is expressed as:

\begin{equation}
{{\cal L}_{\rm{P}}}\left( {{\theta _g}} \right) = \frac{1}{{{h_g}}}\sum\limits_{t = 1}^{{h_g}} {{\rm{ReLU}}} \left( {{\Delta ^{t + 1 \to t}}} \right) \quad {\rm{    for     }} \quad t \in [1,...,{h_g}].
\end{equation}

\subsection{System health state parameter inference with surrogate model}
The evolution of the system health condition can be described by system health indicators that for complex systems are often not directly observable or measurable and that fulfil certain criteria, such as for example a monotonous decrease over operating time (following the assumption that a system cannot repair itself without an external maintenance action \cite{arias2021aircraft}). Any synthetically generated condition monitoring data needs, therefore, to comply with the underlying degradation processes, including the requirement that the health indicators of the generated synthetic data are monotonically decreasing. As mentioned above, health indicators can typically not be directly observed for complex systems. Therefore, in our case, they need to be inferred from a system performance model that is based on the sensor measurements (${x_s}$) and the corresponding scenario-descriptor operation conditions ($w$). However, the physics-based system performance model is not available. DL-based models can describe this complex inherent relationship well, which gives the opportunity to use DL-based surrogate model to infer health parameters. To enable the inference of health indicators in real time, we propose to develop a surrogate model to obtain the inferred $\hat \theta $. A basic deep neural network (DNN) with three layers, ${{\cal F}_{{\rm{DL}}}}$, is developed as a surrogate model to learn the relationship between real $[w,{x_s}]$ and $\theta $, and output the inferred $\hat \theta $.

As mentioned above, given the scenario-descriptor ($w$) and the observable sensor measurements (${x_s}$), ${{\cal D}_{{\rm{raw}}}} = \{ {W_i},{X_{{s_i}}}\} _{i = 1}^U$ are provided as input to the CPI-GAN which generates the synthetic TTF trajectories, ${{\cal D}_{{\rm{syn}}}} = \{ {W_g},{X_{{s_g}}}\} _{g = 1}^G$. $G$ denotes the number of synthetic units, which is specified by the user. In order to ensure that the TTF trajectory follows realistic degradation processes, i.e., that the health indicators corresponding to the generated trajectories meet the monotonicity requirement, regularizations (e.g., the physics-informed loss function) need to be applied during the generation process for health parameters ($\theta $) that are not monotonous. In CPI-GAN, for synthetic trajectories $[{w_g}^t,{x_{{s_g}}}^t]$ generated in each time step, the trained surrogate model is used to predict the inferred health parameters ${\hat \theta _g}^t$, which corresponds to the synthetic trajectories. Once the inferred health indicators are obtained, the proposed physics-informed loss function is used as an additional regularization term in CPI-GAN to impose a penalty on the part of the generation process to ensure that the TTF trajectories follow realistic degradation processes of the impacted components.

\subsection{Deep learning-based prognostics model}

To evaluate the performance of the proposed framework, we compare it to a purely data-driven approach. The performance of the prediction model is not the focus of this study. Deep CNN architectures have demonstrated an excellent performance on time-series for predictive maintenance in recent works \cite{chao2022fusing, mo2021variational}. To enable a fair comparison between the proposed framework and the other method that have been applied in the same dataset, the same one-dimensional convolutional neural network (1D-CNN) architecture from \cite{chao2022fusing} is applied as the prediction model. 

\begin{figure}
  \centering
  \includegraphics[scale=0.53]{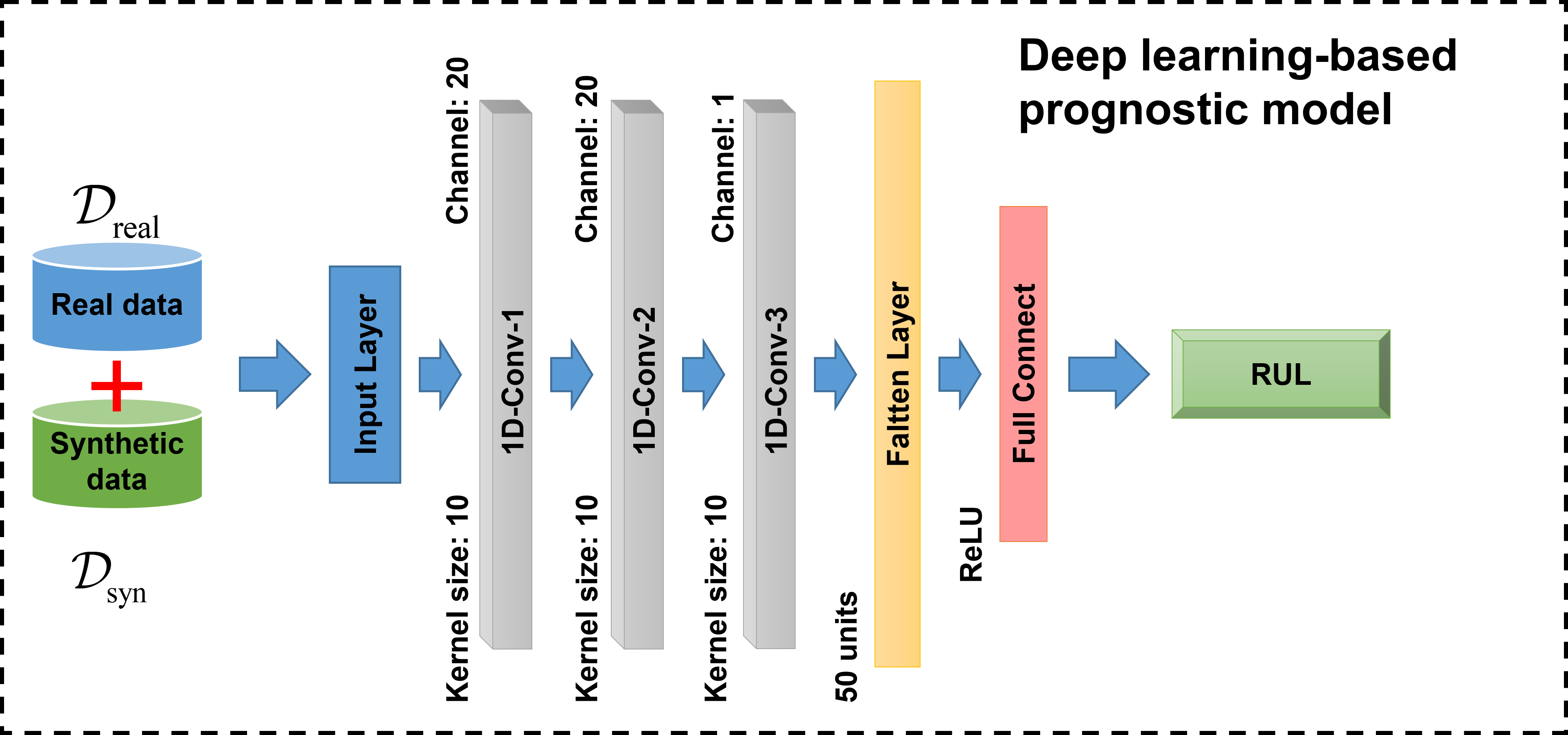}
  \caption{The structure of the 1D CNN.}
  \label{fig:fig4}
\end{figure}

In this part, 1D-CNN based models are used as the baseline approach. The same architecture is also used in the proposed hybrid framework. The structure of the 1D-CNN model is shown in Figure \ref{fig:fig4}. The architecture of the 1D-CNN used in this study comprises six layers, which is the same as the architecture proposed in \cite{chao2022fusing}. The network has three convolutional layers with filters of size ten. The first two convolutions have twenty channels, and the third convolution has only one channel. Zero padding is used to keep the feature map through the network. The resulting 2D feature map is flattened and the network ends with a 50-way fully connected layer followed by a linear output neuron. The network uses ReLU as the activation function in all layers.

\section{Case study}
\label{sec:Case study}
In this section, the dataset description, the procedures of data pre-processing and the detailed parameters of the proposed framework (both generation model and prognostics model) are presented.

\subsection{Dataset description}
\begin{figure}
  \centering
  \includegraphics[scale=0.4]{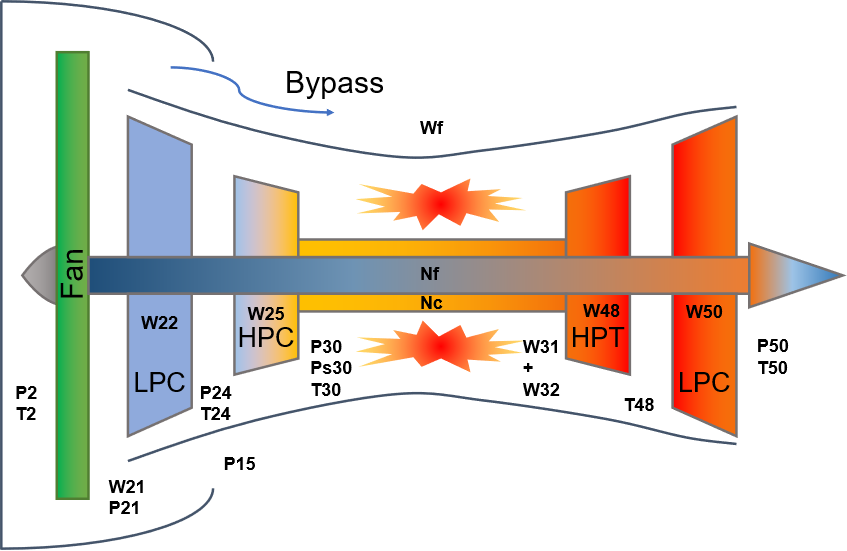}
  \caption{The structure and the sensor location of the turbofan engine (modified from \cite{raissi2019physics}).}
  \label{fig:fig5}
\end{figure}

In this work, the N-CMAPSS \cite{arias2021aircraft} dataset is used to test our proposed hybrid framework for RUL prediction. The N-CMAPSS dataset offers high fidelity TTF degradation trajectories of turbofan engines. It overcomes some of the shortcomings of the old CMAPSS dataset by incorporating real recorded flight conditions and extending the underlying degradation model by relating the degradation process to its operation history \cite{arias2021aircraft}. On overview of the structure the turbofan engine is shown in Figure \ref{fig:fig5}. Table~\ref{tab:table1}, Table~\ref{tab:table2} and Table~\ref{tab:table3} give an overview of the datasets with the scenario descriptors (Table~\ref{tab:table1}), the model health parameters (Table~\ref{tab:table2}), and the sensor measurements (Table~\ref{tab:table3}) in the dataset.

\begin{table}
 \caption{Details of the scenario descriptors ($w$) \cite{arias2021aircraft}.}
  \centering
  \begin{tabular}{lll}
    \toprule

    Symbol     & Description     & Units  \\
    \midrule
    alt      & Altitude                        & ft     \\
    Mach     & Flight Mach number              & -      \\
    TRA      & Throttle–resolver angle         & \%  \\
    T2       & Total temperature at fan inlet  & °R  \\
    \bottomrule
  \end{tabular}
  \label{tab:table1}
\end{table}

\begin{table}
 \caption{Details of the model health parameters ($\theta$) \cite{arias2021aircraft}.}
  \centering
  \begin{tabular}{lll}
    \toprule

    Symbol     & Description     & Units  \\
    \midrule
    HPT$\_$eff$\_$mod      & HPT efficiency modifier    & -     \\
    LPT$\_$eff$\_$mod      & LPT efficiency modifier    & -      \\
    LPT$\_$flow$\_$mod     & HPT flow modifier          & -  \\
    \bottomrule
  \end{tabular}
  \label{tab:table2}
\end{table}

\begin{table}
 \caption{Details of the sensor measurements (${x_s}$) \cite{arias2021aircraft}.}
  \centering
  \begin{tabular}{llllll}
    \toprule

    Symbol     & Description     & Units  & Symbol     & Description     & Units\\
    \midrule
    T24     & Total temperature at LPC outlet     & °R     & P24     & Total pressure at LPC outlet        & psia  \\
    T30     & Total temperature at HPC outlet     & °R     & Ps30    & Static pressure at HPC outlet       & psia   \\
    T48     & Total temperature at HPT outlet     & °R     & P40     & Total pressure at burner outlet     & psia  \\
    T50     & Total temperature at LPT outlet     & °R     & P50     & Total pressure at LPT outlet        & psia  \\
    P15     & Total pressure in bypass-duct       & psia   & Wf      & Fuel flow                           & pps  \\
    P2      & Total pressure at fan inlet         & psia   & Nf      & Physical fan speed                  & rpm  \\
    P21     & Total pressure at fan outlet        & psia   & Nc      & Physical core speed                 & rpm  \\
    \bottomrule
  \end{tabular}
  \label{tab:table3}
\end{table}

Similar as in previous research \cite{chao2022fusing}, the proposed framework is demonstrated and evaluated on DS-02, which is a sub dataset in the N-CMAPSS dataset. There are six training units ($u = 2, \enspace 5, \enspace 10, \enspace 16, \enspace 18, \enspace 20$) and three test units ($u = 11, \enspace 14, \enspace 15$) in DS-02. The overview of DS-02 is listed in Table~\ref{tab:table4}. As shown in Table~\ref{tab:table4}, the flights are divided into three classes depending on the length of the flight. Each flight is divided into cycles, covering climb, cruise, and descend operations. More details on the dataset can be found in \cite{arias2021aircraft}. 

\begin{table}
 \caption{Dataset DS-02 overview: The unit number ($u$), length of the sensory signal (${m_u}$), flight class ($F{C_u}$), failure mode (FM), average flight time of cycle (AFT) and end-of-life time (${t_{EOL}}$) of each unit within the training and testing datasets.}
  \centering
  \begin{tabular}{llllll}
    \toprule
    \multicolumn{2}{c}{Training dataset (DS-02)}                   \\
    \cmidrule(r){1-2}
    Unit ($u$)     & ${m_u}$     & FM     & ${t_{EOL}}$    & FC   & AFT ($s$)\\
    \midrule
    2              & 0.85M  & HPT         & 75             & 3 & 11375.23  \\
    5              & 1.03M  & HPT         & 89             & 3 & 11611.46  \\
    10             & 0.95M  & HPT         & 82             & 3 & 11618.43  \\
    16             & 0.77M  & HPT+LPT     & 63             & 3 & 12147.54  \\
    18             & 0.89M  & HPT+LPT     & 71             & 3 & 12545.34  \\
    20             & 0.77M  & HPT+LPT     & 66             & 3 & 11638.79  \\
    \bottomrule
    
    \multicolumn{2}{c}{Testing dataset (DS-02)}                   \\
    \cmidrule(r){1-2}
    Unit ($u$)     & ${m_u}$     & FM     & ${t_{EOL}}$    & FC   & AFT ($s$)\\
    \midrule
    11             & 0.66M  & HPT+LPT     & 59             & 3 & 11245.68  \\
    14             & 0.16M  & HPT+LPT     & 76             & 1 & 2062.87  \\
    15             & 0.43M  & HPT+LPT     & 67             & 2 & 6469.70  \\
    \bottomrule
  \end{tabular}
  \label{tab:table4}
\end{table}

There are two different failure modes in DS-02. Units 2, 5 and 10 have the failure mode of abnormal HPT efficiency decline. Units 16, 18 and 20 are affected by more complex failure modes that affect the efficiency and flow rate of LPT and the efficiency degradation of HPT. The test units are affected by the same two failure modes. The initial deterioration state of each unit is different, and the degradation of system components follows a random process, first linear normal degradation, then steeper abnormal degradation. In DS-02, flight class is another important variable, as shown in Table 4, there are three different flight classes in DS-02. As described in \cite{arias2021aircraft}, different flight classes represent different flight durations. Different flight times produce different degradation trajectories. The degradation trajectories in the training dataset are collected from engines that operated in flight class 3. However, in the testing dataset, the degradation trajectories belong to three different flight classes. The results in the operation scenarios included in the training dataset cannot cover the scenarios in the testing dataset.

\subsection{Data pre-processing and sequence sample generation}
Due to the large size of the N-CMAPSS dataset, we first reduce the size of the dataset to improve the iteration speed of the generating and training progress. Firstly, we adapt the sensor readings format from the double-precision floating-point format to the half-precision floating-point format. Secondly, we reduce the sampling frequency from 1Hz to 0.1Hz by decimation. 

A side effect of reducing the sampling frequency is that we achieve a longer receptive field for the same number of layers in the neural network. A potential disadvantage of these two dataset reduction measures is that information is lost. The motivation behind this is that long ranging trends are more important for prognostics tasks than local perturbations within a single flight cycle.
In the next pre-processing step, both real and synthetic sensor readings are normalized separately within each of the flight classes (FCs):

\begin{equation}
\tilde X_s^{FC} = \frac{{X_s^{FC} - \mu _s^{FC}}}{{\sigma _s^{FC}}}
\end{equation}

where $\tilde X_s^{FC}$ represents the scaled value of the $s{\rm{ - th}}$ sensor within each FC, $X_s^{FC}$ are the original sensor readings, $\mu _s^{FC}$ and $\sigma _s^{FC}$ are the mean and standard deviation of the $s{\rm{ - th}}$ sensor in the units under each FC, respectively. Scaling is performed under each FC.

A sliding time window technique is applied to sequence both training and testing data, this approach is the same as previously applied on the same dataset. Specifically, the first sequence takes samples from time point 1 to time point $l$, the length of second sequence is $2 \sim (l + 1)$, and so on for the whole degradation trajectory, where $l$ is the width of the sliding window. In this study, $l$ is set to 50, stride is set to 1, and the number of input variables is 18. Finally, the dimension of each input sequence is $x \in {R^{1 \times 50 \times 18}}$. For an entire degradation trajectory, there are (${N_U} - 49$) number of sequences is then concatenated to form the batch dimension. The resulting input tensor has the shape $[{N_U} - 49,50,18]$. Each sequence has a RUL target label associated with it.

\subsection{Domain knowledge-based basic physics constraints}
\label{sec:Domain knowledge-based basic physics constraints}
In this part, a statistical analysis is used to analyze all training units in the N-CMAPSS (including all eight sub-datasets, from DS-01 to DS-08) dataset to obtain the range of the scenario descriptors and transition times under different flight classes as the domain knowledge used in the basic physics constraints (in real applications, this range can be defined by domain experts).

\begin{figure}
  \centering
  \includegraphics[scale=0.8]{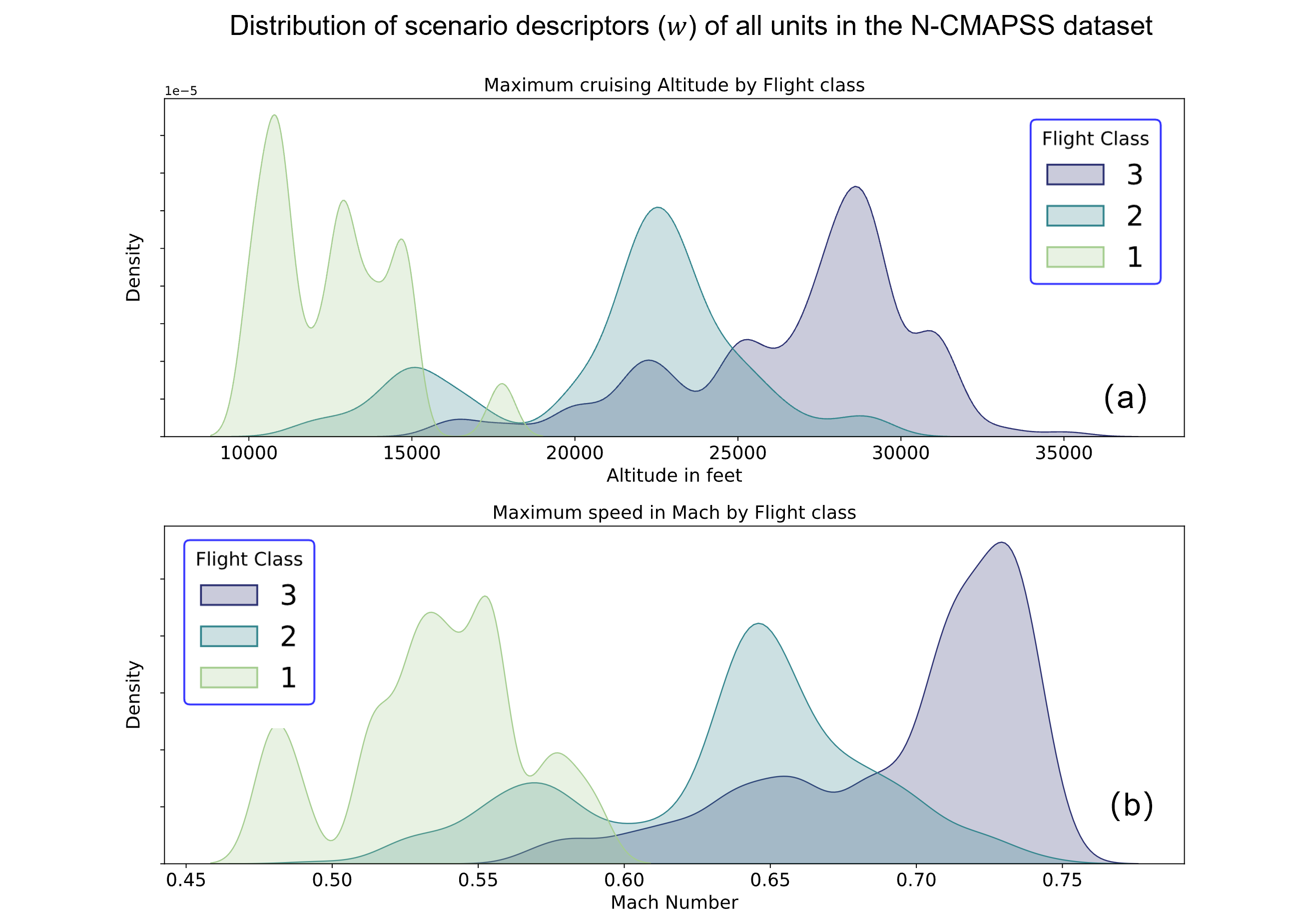}
  \caption{Distributions of three scenario descriptors: (a) the maximum cruising altitude (b) the maximum speed under three flight classes from all eight training datasets in N-CMAPSS.}
  \label{fig:fig6}
\end{figure}

As shown in Table~\ref{tab:table4}, while the training dataset only contains units under FC 3, testing units are covering all the three FC 1, 2 \& 3, and the FC affects the flight time of each cycle in the whole engine degradation trajectory. Similarly, the operating conditions in scenario descriptors $(w)$ depend on FCs. As shown in Figure \ref{fig:fig6}, the maxi-mum cruising altitude and speed of the flight in each cycle are significantly different from each other when they belong to different FCs.

Since the training dataset in DS-02 already contains sufficient units from long flights (FC3), we focus on generating some synthetic units belonging to FC1 and FC2 to improve the prediction accuracy and generalization performance regarding testing dataset. In the study, the range of these variables (i.e., the flight time of each cycle, maximum cruising altitude, maximum speed, etc.) is used as the basic physics constraints for CPI-GAN to generate time to failure trajectories for synthetic units.

\begin{figure}
  \centering
  \includegraphics[scale=0.8]{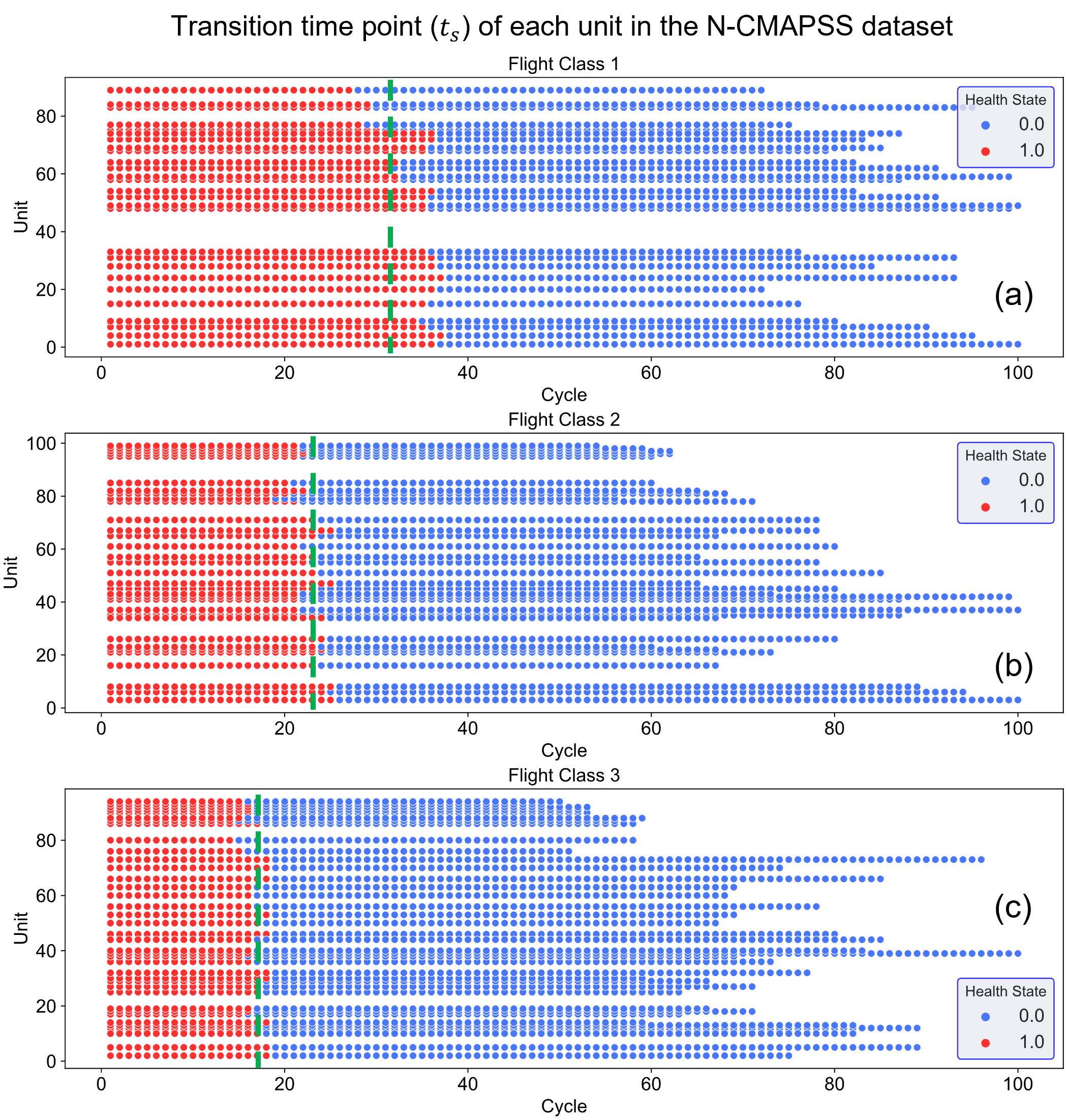}
  \caption{The transition time point of units under three different flight classes in the N-CMAPSS dataset, (a) FC1, (b) FC2, (c) FC3. The horizontal coordinate denotes the number of the unit, and the vertical coordinate denotes the number of the cycle. The red dots represent the engines in the state of normal degradation, and the blue dots represent the units in the state of abnormal degradation. The time when the red dots turn into blue dots is the transition time point. The green dashed lines show the mean value of the transition time point for units belonging to different flight classes.}
  \label{fig:fig7}
\end{figure}

After obtaining the above constraints, we need to obtain the range of the transition time ($t_s^u$). As described in \cite{arias2021aircraft}, the transition time of each unit is another important variable dependent on the FCs. To generate degradation trajectories belonging to FC1 \& FC2, the location of the transition time ($t_s^u$) points is also important. As shown in Figure \ref{fig:fig7}, by analyzing all sub-datasets in N-CMAPSS, it can be found that the locations of transition time for the units belonging to FC1 \& FC2 are significantly different from those belonging to FC3.

Based on the analysis, the basic physics constraints used for data generation in CPI-GAN are summarized in Table~\ref{tab:table5}. In this study, we generate full trajectories for two synthetic units (marked as ${D_{{\rm{S - FC1 - 01}}}}$ and ${D_{{\rm{S - FC1 - 02}}}}$) from FC1 and two synthetic units (marked as ${D_{{\rm{S - FC2 - 01}}}}$ and ${D_{{\rm{S - FC2 - 02}}}}$) from FC2 . Different numbers of synthetic units belonging to different FCs affect the balance of data in the enhanced training dataset (${{\cal D}_{{\rm{real + syn}}}}$). The optimal number of synthetic units is analyzed in section 5.3. The flight times, cruising altitude, and speed are randomly selected from the ranges shown in Table~\ref{tab:table5}. The number of cycles in each synthetic unit is determined by the time before the fault initiation (it is randomly sampled from the range of the transition time point, for FC 1, the range is (27, 38), for FC 2, the range is (20, 25)). The length of the entire trajectory is then determined by the number of cycles after the transition time point (randomly sampled from the range (36, 45). Next, the above basic physics constraints are imposed on CPI-GAN as the initial settings that can be controlled manually to generate four synthetic units belonging to FC1 and FC2.

\begin{table}
 \caption{Summary of the constraints used for generating synthetic units by CPI-GAN.}
 \resizebox{\textwidth}{!}{
  \centering
  \begin{tabular}{llllllll}
    \toprule

    Dataset     & unit     & FC  & Range of flight time      & Cycles     & Position of  & Range of cruising  & Range of \\
     &     &   &  of each cycle (s)     &  (${t_{{\rm{EOL}}}}$) & the ${t_s}$ & altitude (feet) & speed (Mach)\\
     
    \midrule
                     & ${D_{S - Fc1 - 01}}$     & 1     & [1000, 3900]     & 75        & 36        & [10000, 20000]  & [0.3, 0.6]\\
    Synthetic        & ${D_{S - Fc1 - 02}}$     & 1     & [1000, 3900]     & 80        & 37        & [10000, 20000]  & [0.3, 0.6]\\
    dataset              & ${D_{S - Fc2 - 01}}$     & 2     & [2100, 9500]     & 60        & 23        & [10000, 30000]  & [0.5, 0.75]\\
                     & ${D_{S - Fc2 - 02}}$     & 2     & [2100, 9500]     & 65        & 22        & [10000, 30000]  & [0.5, 0.75]\\
    \bottomrule
  \end{tabular}
  }
  \label{tab:table5}
\end{table}

\subsection{Optimal parameters of the framework}
In the generation part, an RNN variant called gated activation unit (GRU) with three layers is selected as the basic architecture of all sub-modules (embedding network, generator, and discriminator) in CPI-GAN. The hidden dimensions four times the size of the input features, and the dimension of the latent space is half that of the input features \cite{yoon2019time}. Grid search is used to obtain the best model parameters, the optimal parameters of the CPI-GAN are listed in Table~\ref{tab:table6}. $\lambda $ and $\mu $ are the coefficients of the supervised loss, details on the supervised loss can be found in \cite{yoon2019time}, here, we choose the same settings as the original TimeGAN. The surrogate system model is developed with an FNN with three hidden layers. We choose FNN as the surrogate model because of its simple structure and its ability to achieve the required inference accuracy. All the three hidden layers have 64 hidden units. ReLU activation function is applied in all hidden layers.

\begin{table}
 \caption{The optimal parameters of CPI-GAN.}
  \centering
  \begin{tabular}{ll}
    \toprule

    Embedding \& Recovery \& Generator \& Discriminator modules       & Hyperparameters     \\
    \midrule
    Module = 'GRU'                                                    & Sequence length: 50     \\
    Activation function of input layer: tanh                          & Training iterations: 50000   \\
    Activation function of output layer: Sigmoid                      & Batch size: 128     \\
    Hidden dimensions: 72                      & $\mu  = 10$  \\
    Number of layers: 3                      & $\lambda  = 1$  \\
    Latent space dimension: 9                      & $\lambda _{{\rm{ ReLU}}} = 0.0002$  \\
    \bottomrule
  \end{tabular}
  \label{tab:table6}
\end{table}

In the prediction part, 1D-CNN model is used as the baseline approach. It is also used in the proposed hybrid framework. The architecture of the 1D-CNN in this study comprises six layers, which is same as the architecture proposed in \cite{chao2022fusing}. The operating conditions and sensors measurements $[{x_s},w]$ are used as the inputs. For 1D-CNN models’ requirements, the original dataset is normalized by Eq. (8) and then sequenced by the sliding time window approach \cite{chao2022fusing}. In the pre-processed dataset, 70\% of the sequences are used for training and 30\% for validation. Once the model is trained, it was used to predict the RUL on the testing set. For hybrid approach, the same 1D-CNN architecture is applied to predict the RUL. Six units from the training dataset and four synthetically generated units are used as inputs.
All of the models used in this research are implemented in Python using the Keras and Tensorflow 2.6. The models are trained for 50 epochs using the Adams optimizer. The optimal hyperparameters are obtained by using grid search. The details of the prediction model’s structure and the optimal hyperparameters are shown in Table~\ref{tab:table7}.

\begin{table}
 \caption{Details of the prediction model structure and the optimal hyperparameters.}
  \centering
  \begin{tabular}{llll}
    \toprule
    \multicolumn{2}{c}{Network structure}                   \\
    \cmidrule(r){1-2}
    Layer (type)     & Output Shape           & Activation    & Parameters \\
    \midrule
    Conv1D$\_$1         & (50, 10)               & ReLU          & 2810  \\
    Conv1D$\_$1         & (50, 10)               & ReLU          & 1010  \\
    Conv1D$\_$1         & (50, 1)                & ReLU          & 101  \\
    Flatten          & (50)                   & -             & 0  \\
    Dense$\_$0          & (50)                   & ReLU          & 2550  \\
    Dense$\_$1          & (1)                    & Linear        & 51  \\
    \bottomrule
    
    \multicolumn{2}{c}{Network hyperparameters}                   \\
    \cmidrule(r){1-2}
    Initial learning rate)     & Dropout rate     & Batch size     & Epoch number  \\
    \midrule
    0.001             & 0.2       & 128             & 50  \\
    \bottomrule
  \end{tabular}
  \label{tab:table7}
\end{table}

\section{Results and discussion}
\label{sec:Results and discussion}
In this section, the results of the synthetic data generation and the prediction results of RUL are analyzed and discussed, and the prediction results are compared the baseline approaches.

\subsection{Analysis of the plausibility of synthetic data}
\begin{figure}
  \centering
  \includegraphics[scale=0.8]{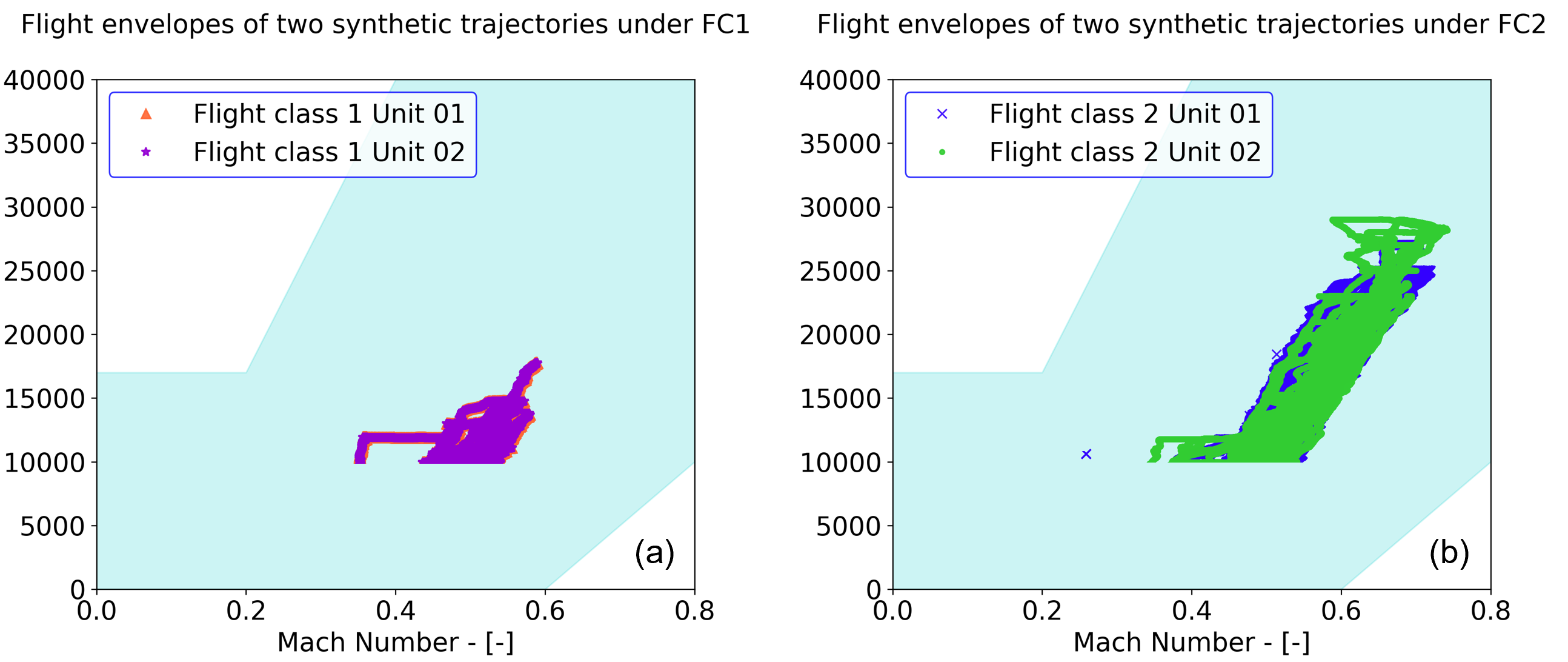}
  \caption{The flight envelopes of two synthetic units belong to FC1 (a), and two synthetic units belong to FC2 (b) are provided according to the readings of flight altitude and speed. The shaded area (light blue) denotes the acceptable operation envelope of the CMAPSS dynamical mode according to the N-CMAPSS documentation \cite{arias2021aircraft}.}
  \label{fig:fig8}
\end{figure}

To verify that the synthetically generated trajectories are consistent with the real trajectories, several evaluations are performed. Figure \ref{fig:fig8} shows the flight envelopes of two synthetic units belong to FC1, and two synthetic units belong to FC2. All the flight envelope recording points are in the acceptable operation envelope area. It indicates that the operation conditions ($w$) generated by CPI-GAN are consistent with the real trajectories and follow the basic physics constraints.

\begin{figure}
  \centering
  \includegraphics[scale=0.8]{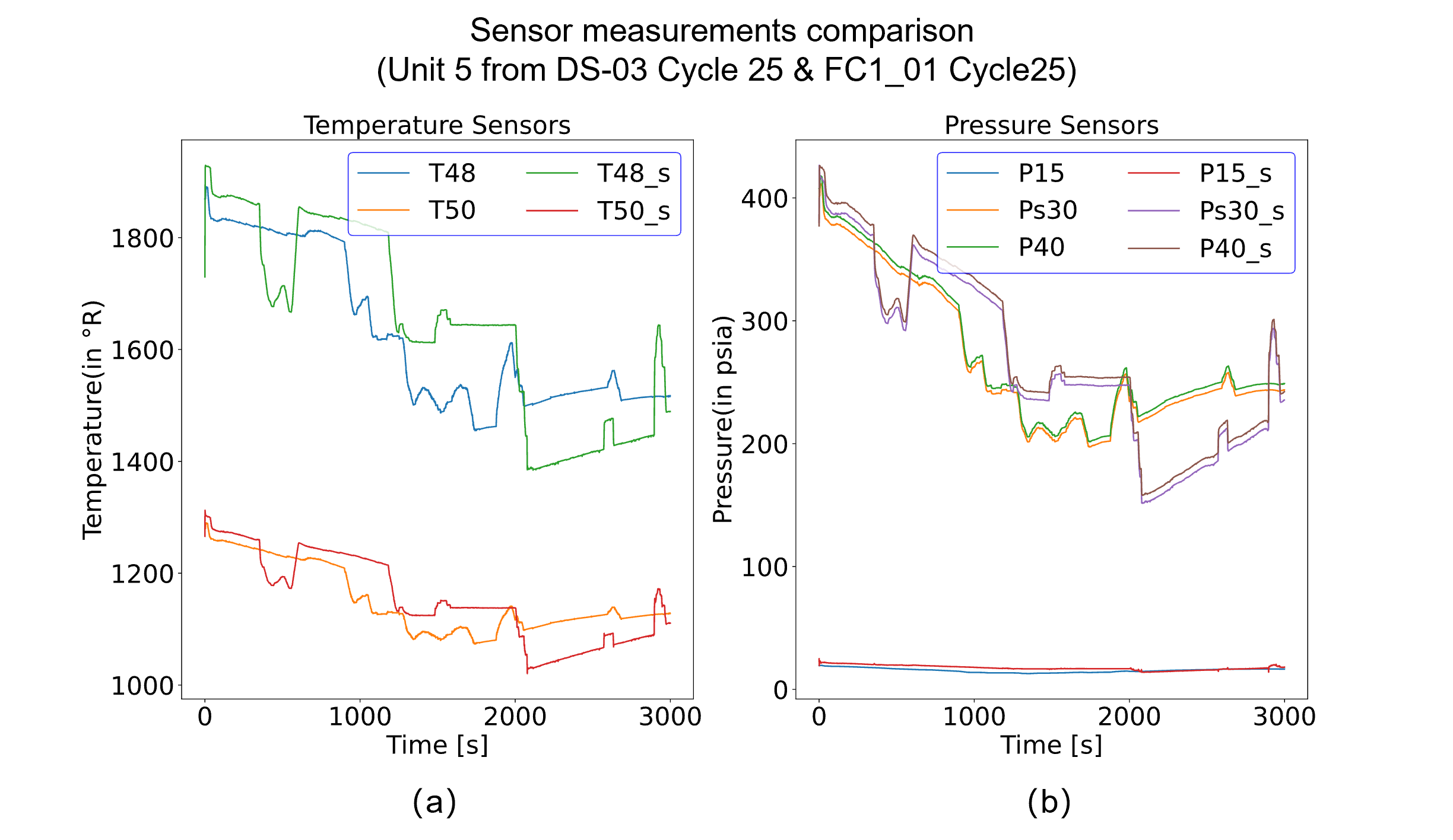}
  \caption{Comparison of two types of sensor measurements from one randomly selected real unit from DS-03 and one synthetic unit (variable names with subscript s represent synthetic data): (a) temperature sensor measurements, (b) pressure sensor measurements.}
  \label{fig:fig9}
\end{figure}

Figure \ref{fig:fig9} shows exemplarily the comparison of sensor measurements (${x_s}$) from one randomly selected real unit from DS-03 and one synthetic unit. The cycle from which the sensor measurements are selected is also randomly selected. DS-03 contains run-to-failure trajectories from FC1 and FC2. However, those units were impacted by a different failure mode. As can be seen in Figure \ref{fig:fig9}, the synthetic trajectory (FC=1) has a certain similarity in the change trend compared to that of the real trajectory (DS-03, FC=1). This shows that the proposed method can generate TTF degradation trajectories that follow realistic operating conditions in the desired flight class (without any training trajectories of the same flight class). 

\begin{figure}
  \centering
  \includegraphics[scale=0.6]{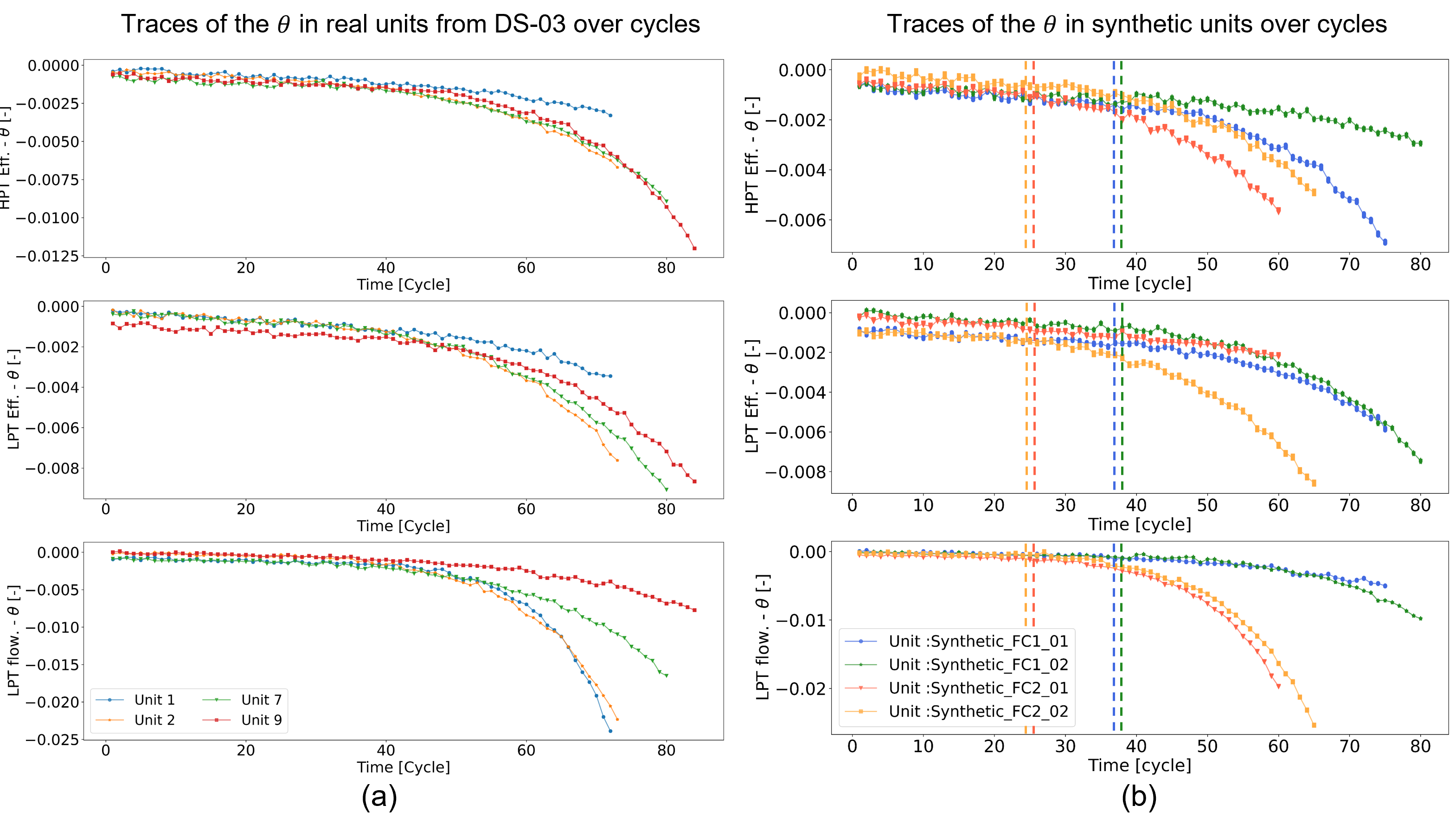}
  \caption{Traces of the degradation on three health parameters (i.e., HPT efficiency, LPT efficiency and LPT flow) over the entire TTF tra-jectory: (a) health parameters of four real units from DS-03, the ${t_{EOL}}$ of these four real units from DS-03 are 72, 73, 80, 84, and (b) health parameters inferred from the synthetic degradation units generated by the CPI-GAN approach. The dashed lines repre-sent the transition time point of each synthetic unit.}
  \label{fig:fig10}
\end{figure}

In addition to evaluating the plausibility of the generated trajectories of the synthetic units from the statistical perspective, we also visualized the traces of health parameters over the entire TTF trajectory. Figure \ref{fig:fig10} shows the health parameters degradation profiles over cycles of four randomly selected units from DS-03 (left column: four selected real units belong to FC1 \& FC2) and the synthetic units (right column). Each line style represents one unit. It can be found that the synthetic units show similar degradation trajectories compared to those of the real data.

By visualizing and comparing the synthetic units and real units from different perspectives (i.e., the flight envelopes, the sensor measurements, the degradation profiles over cycles and times), it shows that the proposed generation approach can ensure that the long-term degradation trends of the synthetic units under different flight classes are plausible to are conform to the realistic operating and degradation conditions. 

As shown in Figure \ref{fig:fig8} and Figure \ref{fig:fig9}, the basic physics constraints ensure that the generated scenario descriptors $w$ are within the limits of the flight envelope. As shown in Figure \ref{fig:fig10}, the physics-informed loss function ensures that the health parameters are satisfying a monotonically decreasing trend, which demonstrates that the generated trajectories for the synthetic units satisfy the underlying degradation laws.

\subsection{Analysis of the remaining useful life prediction results}

The RUL prediction is performed based on the same 1D-CNN architecture for both the baseline approach and the proposed hybrid approach. In this section, the performance of the two approaches is compared based on two performance metrics (i.e., RMSE and $score$). Table 8 shows the RUL prediction performance of both the hybrid approach (trained on the six real units and four synthetic units) and the baseline approach (trained on the six real units). The variables used for training in these two approaches are the same. In the baseline approach, only real data is applied. In the proposed hybrid approach, real data is first used to generate the synthetic trajectories and then, the synthetic data and the real data are combined for training the final RUL prediction model. Both the baseline approach and the proposed hybrid approach are validated on the same testing units (Unit 11, 14, 15).

\begin{table}
 \caption{Overview of the results in terms of RMSE and $score$ of the hybrid framework and the baseline approach on the same testing set (Unit 11, 14, 15), respectively. The Mean and standard deviation of the prediction results are obtained over five runs.}
  \centering
  \begin{tabular}{llll}
    \toprule
    \multicolumn{2}{c}{DS-02}                   \\
    \cmidrule(r){1-2}
    Metric                          & Baseline           & Hybrid    & rel. Delta \\
    \midrule
    RMSE [cycles]                   & 4.64$\pm$0.14               & \textbf{3.82$\pm$0.11}          & -17.67\%  \\
    $score \times {10^5}$ [-]       & 0.48$\pm$0.05               & \textbf{0.23$\pm$0.04}          & -52.08\%  \\
    \bottomrule
  \end{tabular}
  \label{tab:table8}
\end{table}

With a reduction of 17.67\% in RMSE and 52.08\% in $score$, the hybrid approach clearly outperforms the purely data-driven approach with the same 1D-CNN architectures. These improvements are mainly attributed to the addition of four synthetic units belonging to FC1 and FC2. Moreover, since the score penalizes overestimation rather than underestimation, while RMSE is a symmetric metric, the significant reduction of the score metric indicates that the proposed framework handles RUL overestimation more effectively.

\begin{figure}
  \centering
  \includegraphics[scale=0.4]{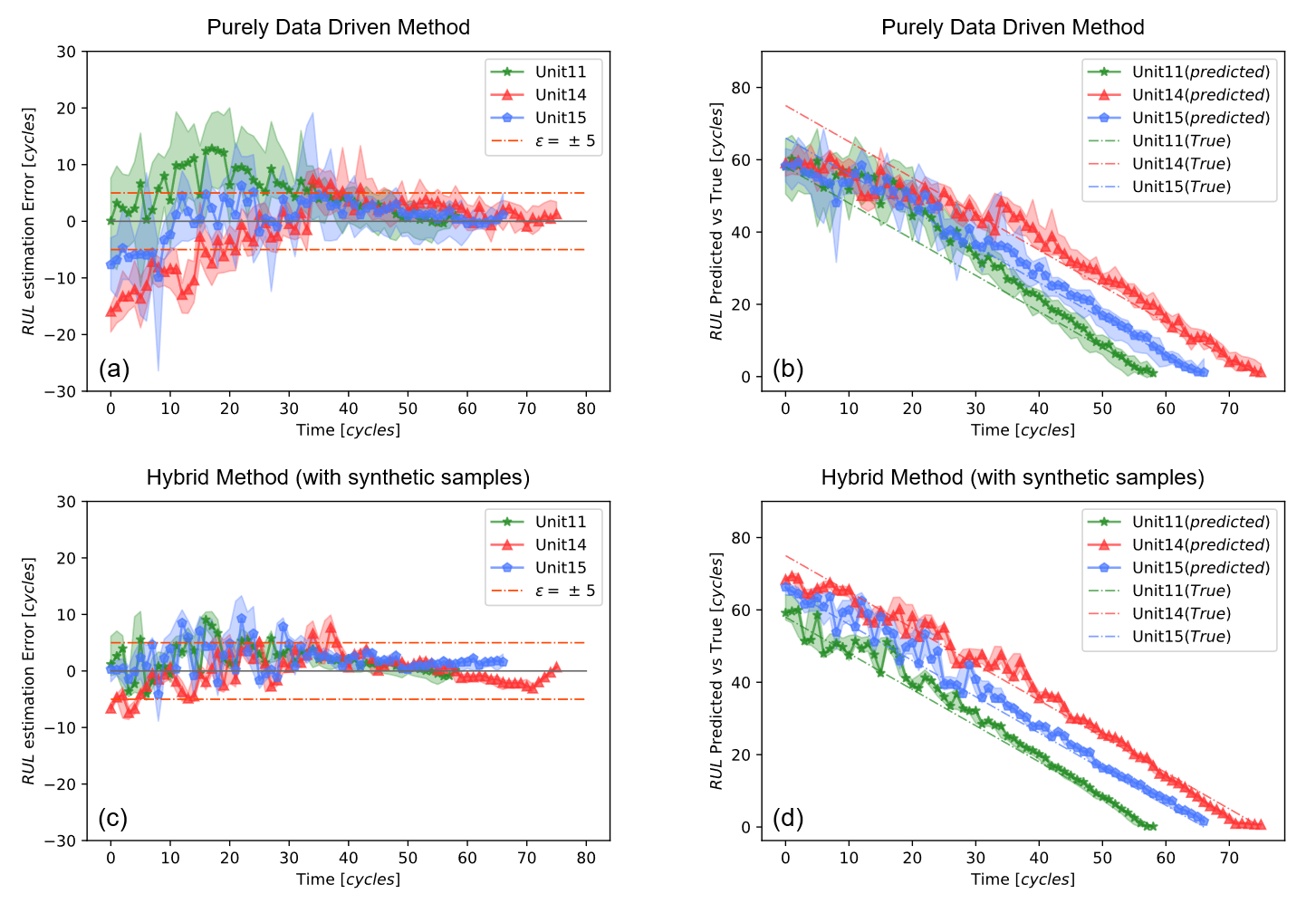}
  \caption{Curves of RUL prediction results (right column) and RUL prediction errors (left column) based on DS-02. The curves of predicted results represent the outputs of the hybrid framework (d) and the purely data-driven approach (b) on Units 11, 14 and 15 successively. The sloping dashed lines represent the true RUL values of each testing unit. The curves of RUL prediction errors of the hybrid framework and the purely data-driven approach on each unit are shown in (a) and (c). The dotted lines are the average RUL prediction errors at each cycle, while the shaded surface represents the uncertainty bounds for RUL predictions within each cycle. The orange dashed horizontal lines correspond to ± 5 cycles error bars. The three test units are shown: Unit 11 (green), Unit 14 (red), and Unit 15 (blue).}
  \label{fig:fig11}
\end{figure}

The improvement of the prognostic performance is also observed on all of the individual test units. Figure \ref{fig:fig11} (left column) shows the prediction errors between true RUL and predicted RUL of the proposed hybrid framework (bottom) and the purely data-driven method (top) for each of the testing units. RUL prediction under the purely data-driven approach (baseline) at any cycle has a large variation compared to that of the proposed framework. It can be observed that the purely data-driven approach performs poorly in the early stage of the degradation. In contrast, the proposed hybrid approach shows a higher prediction accuracy throughout the entire degradation cycle. It demonstrates that the proposed framework can effectively improve the RUL prediction accuracy.

Figure \ref{fig:fig11} shows a comparison of true RUL values and the predicted RUL values (right column) inferred by the purely data-driven method (top) and the proposed hybrid framework (bottom) for each of the testing units. The shaded area shows the variability of the RUL predictions within each cycle. Compared to the purely data-driven method, the prediction results of the hybrid framework show a smaller variation and bias, especially for Unit 14 and Unit 15 (which are from FC 1 and FC2). It indicates that the addition of synthetic units belonging to FC1 and FC2 can effectively improve the generalization performance and the accuracy of the prediction model.

\subsection{Impact of the number of generated synthetic units}
The results shown in the previous section indicate that the proposed hybrid framework outperforms the baseline method (while using the same 1D-CNN architecture). These results support the commonly known hypothesis that the better the data quality and representativeness of the training dataset, the better the performance. Representativeness becomes particularly important if there is a domain gap between the training and the testing data. In the considered case study, all six units in the training data belong to FC3. However, the three units for testing belong to three different flight classes, resulting in a domain gap between the training and the testing dataset. In this research we focus on generative approaches to address the challenge of a domain gap between the training and the testing dataset.

To evaluate how the number of the generated synthetic units impacts the prediction results, we perform four experiments with a different number of synthetic units. For each experiment, half of the synthetic units belongs to FC1, and the other half belongs to FC2. Note that all the synthetic units and the units from the testing set share the same failure mode. The effect of the failure mode of the synthetic units on the prediction results is evaluated in detail in the next section. The prognostics performance of the baseline approach and the hybrid framework trained with a different number of synthetic units is shown in Table~\ref{tab:table9}. As can be seen from Table~\ref{tab:table9}, the best prediction results are obtained when using four synthetic units (two from FC1 and two from FC2). And important observation is that the proposed hybrid approach with different numbers of synthetic units achieves an improvement in prediction accuracy compared to the purely data-driven approach. Another point worth noting is that too many synthetic units and too few synthetic units both yield only a small increase in prediction accuracy. It is mainly because there is only one unit (Unit 11) in the testing set which belongs to FC3, while six units in the training set belong to FC3. Excessive generation of units belonging to FC1 or FC2 can effectively improve the prediction accuracy on Unit 14 (FC1) and Unit 15 (FC2), but it reduces the prediction accuracy on Unit 11(FC3), resulting in a decline of prediction performance on the entire test set. In contrast, too few synthetic units belonging to FC1 or FC2 are not sufficient for the prediction model to learn the degradation information from units belonging to FC1 and FC2. Therefore, four synthetic units are used as the optimal number of units in this study.

\begin{table}
 \caption{Overview of the RMSE and $score$ metrics using the proposed hybrid framework and the baseline method on testing set (Unit 11, 14, 15). Four different numbers of synthetic units are used to train the 1D-CNN model. Mean and standard deviation of the pre-diction results are obtained over five runs.}
 \resizebox{\textwidth}{!}{
  \centering
  \begin{tabular}{llllll}
    \toprule

    Number of    & Eight synthetic units & Six synthetic units & Four synthetic units & Two synthetic units  & No synthetic units  \\
    training units   & \& Six real units  &  \& Six real units   &  \& Six real units  &  \& Six real units  &  \& Six real units \\
    \midrule
    Method                 & Hybrid     & Hybrid     & Hybrid     & Hybrid        & Baseline  \\
    RMSE [cycles]    & 4.42$\pm$0.16     & 4.08$\pm$0.21     & \textbf{3.82$\pm$0.11}     & 4.56$\pm$0.16        & 4.64$\pm$0.14 \\
    $score \times {10^5}$ [-]                 & 0.45$\pm$0.11     & 0.33$\pm$0.05     & \textbf{0.23$\pm$0.04}     & 0.51$\pm$0.12        & 0.48$\pm$0.05 \\

    \bottomrule
  \end{tabular}
  }
  \label{tab:table9}
\end{table}

\subsection{Impact of the failure modes in synthetic units}
In addition to the number of synthetic units, another factor that affects the performance of proposed framework are the failure modes of the generated units. As shown in Table~\ref{tab:table1}, there are two failure modes (FMs) in the real training set. Unit 2, Unit 5, and Unit 10 are subject to FM 1 (HPT efficiency degradation). For Unit 16, Unit 18, and Unit 20, they are subject to FM 2 (HPT efficiency, LPT efficiency and LPT flow degradation). However, in the testing dataset, all three units are subject to FM 2. Accordingly, four synthetic units in the training set are set to be affected by FM 2. In this way, the synthetic data is specifically generated for the failure mode present in the test dataset. We assume that once the fault is detected, we are also able to determine the fault type and then apply a targeted data generation. This is a realistic assumption.

In this section, this assumption is relaxed, and data is generated mainly based on the knowledge of training set and some available domain knowledge on the expected fault types (e.g., the results of statistical analysis based on the eight sub-datasets in N-CMAPSS). For this evaluation, four synthetic units are generated, with two synthetic units under FC1 belonging to FM 1 and FM 2, respectively. The other two synthetic units under FC2 be-longing to FM 1 and FM 2, respectively. The prognostics performance of the hybrid framework with the knowledge of the failure modes and the hybrid framework without any knowledge on the failure modes in the testing dataset is shown in Table~\ref{tab:table10}. It can be seen from Table~\ref{tab:table10}, that the addition of synthetic units with the knowledge of the relevant failure modes resulted in a significant improvement in the prediction accuracy of RUL. In fact, we can build specialized prognostics models for each failure mode. However, one of the goals could be to have a generic model for all failure modes. The relaxation of the assumption on the failure mode allows the proposed framework to be adapted to more realistic scenarios, as the true operating conditions and failure modes of the testing units may be difficult to know in advance in practical applications.

\begin{table}
 \caption{Overview of the RMSE and $score$ metrics using the proposed hybrid framework with prior knowledge and the hybrid framework without prior knowledge. Mean and standard deviation of the prediction results are obtained over five runs.}

  \centering
  \begin{tabular}{llll}
    \toprule

    Prior knowledge   & Available  & Unavailable   & rel. Delta \\
    \midrule
    RMSE [cycles]                             & \textbf{3.82$\pm$0.11}     & 4.09$\pm$0.21     &   +7.07\%  \\
    $score \times {10^5}$ [-]                 & \textbf{0.23$\pm$0.04}     & 0.28$\pm$0.09     &   +21.74\%  \\

    \bottomrule
  \end{tabular}

  \label{tab:table10}
\end{table}

\subsection{Sensitivity analysis of penalty threshold}
\label{sec:Sensitivity analysis of penalty threshold}
\begin{figure}
  \centering
  \includegraphics[scale=0.4]{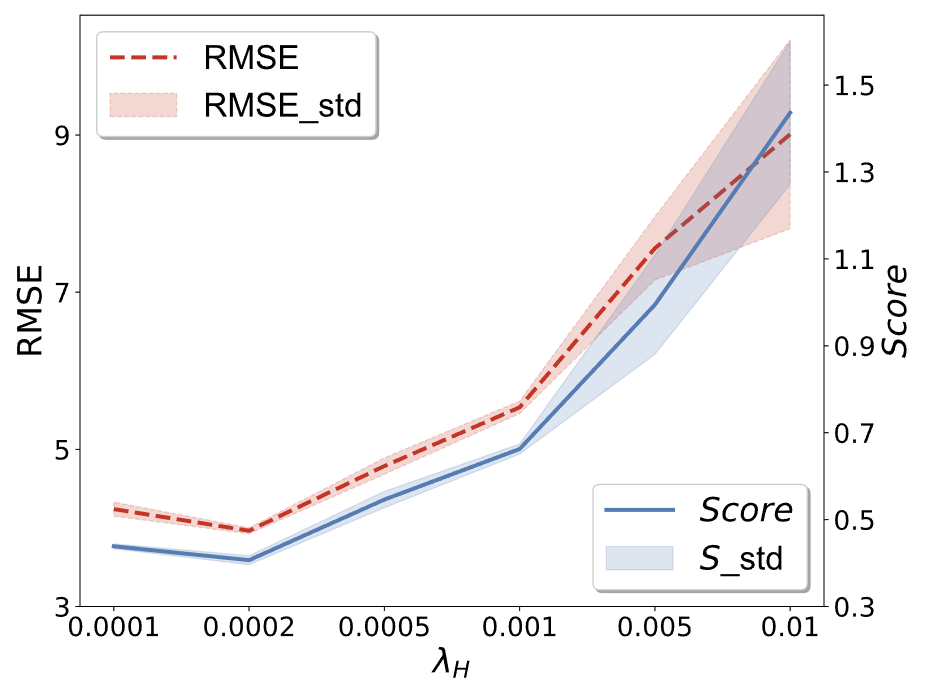}
  \caption{Influence of the penalty threshold on the prediction performance, considered on the DS02 as an example. The red dashed line represents the RMSE values of each penalty threshold ${\lambda _{{\rm{ReLU}}}}$. The blue line represents the score values of each penalty threshold ${\lambda _{{\rm{ReLU}}}}$. The shaded surfaces represent the uncertainty bounds for RMSE and score within each ${\lambda _{{\rm{ReLU}}}}$. Mean and standard devia-tion of the prediction results are obtained over five runs.}
  \label{fig:fig12}
\end{figure}

The penalty threshold is a key parameter in the CPI-GAN and has a significant influence on the quality of the synthetic TTF trajectories and the RUL prediction accuracy. If the penalty threshold is too small, it is not in line with the actual situation of engine degradation. If the threshold is too large, the synthetic data would fluctuate substantially, and the quality of the generated data would also be affected. According to the range of real health state parameters, experiments are conducted with different values for penalty threshold. The results are shown in Figure \ref{fig:fig12}. The penalty threshold values of 0.0001, 0.0002, 0.0005, 0.001, 0.005, and 0.01 are evaluated in the experiments.

According to Figure \ref{fig:fig12}, it can be found that as the penalty threshold increases, both RMSE and the score show a trend of first decreasing and then increasing. When the threshold is set to ${\lambda _{{\rm{ReLU}}}} \ge 0.001$, the variations of the RMSE and score also gradually become larger. The proposed RUL prediction model obtains the best results (i.e., the smallest RMSE, Score and variance) when the penalty threshold is ${\lambda _{{\rm{ReLU}}}} = 0.0002$.

\subsection{Comparison to previous research}
The performance of the proposed framework is compared to a previously published study [9]. In Arias Chao’s et al. study [9], a physics-based performance model was proposed to infer unobservable model parameters related to a system’s components health by solving a calibration problem. These parameters are subsequently combined with sensor readings and used as input to a deep neural network (e.g., the same 1D-CNN architecture as used in this paper), thereby generating a data-driven prognostics model with physics-augmented features. The difference between Arias Chao’s et al. [9] method and the proposed method is that the former used a physics-based model to infer calibration parameters and virtual sensors and add these variables to the original data. In this research, we use the calibration parameters during training as physics-informed loss to impose a realistic degradation trajectory. The results are displayed in Table~\ref{tab:table11}. Compared to the results reported in [9], the per-formance of the proposed framework improves RMSE by 7.73\% and the score by 47.72\%. In addition to these two performance-metrics, the prediction horizon with a prediction error below 5 cycles (${H_{\left| {{\Delta _y}} \right| \le 5}}$) is also used to evaluate the performance of the proposed framework, the details of this performance metric can be found in [9]. Table~\ref{tab:table12} shows the prediction horizon results for each unit and the average value from the proposed framework and the existing method. Under this metric, the proposed framework provides almost identical results on the prediction horizon while maintaining similar prediction accuracy.

\begin{table}
 \caption{RMSE and $s-score$ results comparison.}

  \centering
  \begin{tabular}{lll}
    \toprule

    Method   & RMSE [cycles]   & $score \times {10^5}$ [-] \\
    \midrule
    Arias Chao’s \textit{et al.} best \cite{chao2022fusing}  & 4.14$\pm$0.09     & 0.44$\pm$0.02  \\
    Proposed                                & \textbf{3.82$\pm$0.11}     & 0.23$\pm$0.04    \\
    rel. Delta                               & -7.73\%     & -47.72\%    \\
    \bottomrule
  \end{tabular}

  \label{tab:table11}
\end{table}

\begin{table}
 \caption{Prediction horizon [cycles] comparison for ${\Delta _y} \le 5$ (i.e., ${H_{\left| {{\Delta _y}} \right| \le 5}}$) between existing method and proposed hybrid framework.}

  \centering
  \begin{tabular}{llll}
    \toprule

    Unit number ($u$)  &  Arias Chao’s \textit{et al.} method \cite{chao2022fusing}  & Hybrid framework   & rel. Delta \\
    \midrule
    11                & 31     & \textbf{32}     &   +3\%  \\
    14                & \textbf{43}     & 39     &   -9\%  \\
    15                & 37     & \textbf{40}     &   +8\%  \\
    \midrule
    Avg.              & 37     & 37     &   -\\
    \bottomrule
  \end{tabular}

  \label{tab:table12}
\end{table}

\section{Conclusions}
\label{sec:Conclusions}
Traditional data-driven RUL prediction approaches require large amounts of representative time-to-failure trajectories for training, which are usually not available in practice. One of the main difficulties is that system operating conditions and failure modes to which the collected data belongs do not always cover all relevant scenarios and may, therefore, not be representative for the applied testing dataset. In this paper, we propose a controlled data generation technique based on physical information to generate synthetic data belonging to the unseen scenarios to tackle on the one hand the challenge of limited representativeness and on the other hand the challenge of missing time-to-failure trajectories The proposed approach allows prior knowledge as well as underlying physical laws to generate time-to-failure data under different operating conditions that comply with the underlying physical laws and basic physics constraints. The synthetic time-to-failure data and the real data are then used to train a DL-based model to predict the RUL. We demonstrate in extensive evaluations that the synthetic data is conform with the physical constraints and the degradation trend of the system. The proposed framework is evaluated on the N-CMAPSS dataset, and the proposed framework is able to significantly outperform the purely data-driven approach.

Future research could extend the framework to a more pro-active data generation setup, where the prediction uncertainty along the degradation trajectory for example is used to target the generation of samples in this specific part of the degradation trajectory. Moreover, it would be also interesting to extend the proposed framework to other application cases. 

\section*{Declaration of Competing Interest}
The authors declare that they have no known competing financial interests or personal relationships that could have appeared to influence the work reported in this paper.

\section*{Acknowledgments}
This work was supported by the National Natural Science Foundation of China (Nos. 71931006 and 72101116), the Natural Science Foundation of Jiangsu Province (No. BK20210317).

\bibliographystyle{unsrt}  
\bibliography{references}

\end{document}